%% file: main.tex
\documentclass[runningheads]{llncs}

\usepackage{eccv}

\usepackage{eccvabbrv}

\usepackage{graphicx}
\usepackage{booktabs}
\usepackage{amsmath}
\usepackage{wrapfig}
\usepackage{algorithm}
\usepackage{algpseudocode}
\usepackage{multirow}
\usepackage{subcaption}
\usepackage{comment}

\usepackage[accsupp]{axessibility}  

\usepackage[pagebackref,breaklinks,colorlinks,citecolor=eccvblue]{hyperref}

\usepackage{orcidlink}

\input{shortcuts}

\begin{document}
\emergencystretch 3em

\title{Differentiable Product Quantization for Memory Efficient Camera Relocalization} 

\titlerunning{DPQ for Memory Efficient Camera Relocalization}

\newcommand{\equalcontrib}{\textsuperscript{*}}

\author{Zakaria Laskar\equalcontrib\inst{1} \and
Iaroslav Melekhov\equalcontrib\inst{2} \and
Assia Benbihi\inst{4}\and \\
Shuzhe Wang \inst{2} \and Juho Kannala\inst{2,3} \\
\medskip
$^1$VRG, FEE, Czech Technical University in Prague ~ $^2$Aalto University \\ $^3$University of Oulu ~ $^4$CIIRC, Czech Technical University in Prague}

\authorrunning{Z. Laskar et al.}
\institute{}

\maketitle
\def\thefootnote{*}\footnotetext{denotes equal contribution. \\ Correspondence to: \email{zakaria.nits@gmail.com}, \email{iaroslav.melekhov@aalto.fi}}

\begin{abstract}
  Camera relocalization relies on 3D models of the scene with a large memory footprint that is incompatible with the memory budget of several applications.
  One solution to reduce the scene memory size is map compression by removing certain 3D points and descriptor quantization. This achieves high compression but leads to performance drop due to information loss. To address the memory performance trade-off, we train a light-weight scene-specific auto-encoder network that performs descriptor quantization-dequantization in an end-to-end differentiable manner updating both product quantization centroids and network parameters through back-propagation. In addition to optimizing the network for descriptor reconstruction, we encourage it to preserve the descriptor-matching performance with margin-based metric loss functions.
  Results show that for a local descriptor memory of only 1MB, the synergistic combination of the proposed network and map compression achieves the best performance on the Aachen Day-Night compared to existing compression methods. Our code will be publicly available at \url{https://github.com/AaltoVision/dpqed}.
\end{abstract}

\input{sections/intro}
\input{sections/related}
\input{sections/preliminaries}
\input{sections/method}
\input{sections/experiments}
\input{sections/conclusion}

\PAR{Acknowledgments}
This work was supported by Programme Johannes Amos Comenius CZ.02.01.01/00/22\_010/0003405 and the Czech Science Foundation (GACR) EXPRO (grant no. 23-07973X). JK acknowledges funding from the Academy of Finland (grant No. 327911, 353138) and support by the Wallenberg AI, Autonomous Systems and Software Program (WASP) funded by the Knut and Allice Wallenberg Foundation. We acknowledge OP VVV funded project CZ.02.1.01/ 0.0/0.0/16\_019/0000765 ``Research Center for Informatics'', CSC -- IT Center for Science, Finland, and the Aalto Science-IT project for computational resources. 

\bibliographystyle{splncs04}
\bibliography{main}

\clearpage
\appendix

\input{supplementary/intro}
\input{supplementary/implementation}
\input{supplementary/ablation}

\input{supplementary/results}
\input{supplementary/impact}

\end{document}

%% file: shortcuts.tex
\newcommand{\boldparagraph}[1]{\noindent{\bf #1} }

\DeclareMathOperator*{\argmin}{argmin}

\newcommand{\ba}{\mathbf{a}}

\newcommand{\bc}{\mathbf{c}}\newcommand{\bC}{\mathbf{C}}
\newcommand{\bd}{\mathbf{d}}
\newcommand{\be}{\mathbf{e}}

\newcommand{\bh}{\mathbf{h}}

\newcommand{\bq}{\mathbf{q}}
\newcommand{\bR}{\mathbf{R}}

\newcommand{\bt}{\mathbf{t}}

\newcommand{\bx}{\mathbf{x}}

\newcommand{\nR}{\mathbb{R}}

\usepackage[dvipsnames]{xcolor}
\usepackage{comment}

\newcommand{\PAR}[1]{\vskip4pt \noindent{\bf #1~}}

\usepackage[acronym]{glossaries}
\makeglossaries{}
\newacronym{tab}{Tab.}{Table}
\newacronym{pq}{PQ}{Product Quantization}

\newcommand{\figref}[1]{Fig.~\ref{#1}}
\newcommand{\secref}[1]{Section~\ref{#1}}
\newcommand{\eqnref}[1]{Eq.~\eqref{#1}}
\newcommand{\tabref}[1]{Table~\ref{#1}}

%% file: sections/intro.tex
\section{Introduction}\label{sec:intro}
The development of small and affordable cameras makes visual localization, \ie, the estimation of the position and orientation of a camera from images only, an effective solution for the localization of autonomous systems and mobile devices in practical scenarios.
Applications that use such localization capability include Structure-From-Motion (SfM)~\cite{schoenberger2016sfm}, Simultaneous Localization and Mapping (SLAM)~\cite{cummins2010fab}, and more recently, Augmented and Virtual Reality (AR / VR)~\cite{guzov2021human,sarlin2022lamar}, \eg, Meta Aria glasses, Microsoft Hololens or Apple Vision Pro.

Among existing localization methods, structure-based ones~\cite{sattler2018benchmarking,sarlin2019coarse} set the state-of-the-art:
the scene is represented by a map made of 3D points associated with high-dimensional descriptors~\cite{detone2018superpoint,dusmanu2019d2,revaud2019r2d2}.
A query image is localized by extracting descriptors at keypoint locations in the image and matching these descriptors against those from the map~\cite{sarlin2020superglue, edstedt2023dkm, edstedt2023roma}. 
The resulting 2D-3D matches between the keypoints and the map points are then used to estimate the image's camera pose using geometric solvers integrated into a robust estimation framework~\cite{chum2003locally, fischler1981random, barath2020magsac++}.

Accurate 2D-3D matching is central to the performance of visual localization and is typically enabled by the high-dimensional descriptors~\cite{detone2018superpoint,revaud2019r2d2} in the map.
However, these descriptors increase the memory complexity of the 3D map, limiting the scalability of localization for devices with low memory or low bandwidth. 
This motivates scene compression schemes to reduce the map's size to meet the low memory requirements of industrial applications while preserving localization performance.
Previous works on scene compression follow two main strategies: reduce the set of 3D points to a minimal set 
sufficient for localization (\textit{map compression})~\cite{li2010location,soo20133d,6909460,dymczyk2015keep,7139575,cheng2016data,mera2020efficient} or quantize the feature descriptors (\textit{descriptor compression})~\cite{lynen2015get,7410600}. 
The two strategies can be combined into hybrid methods~\cite{camposeco2019hybrid, Yang_2022_CVPR} to satisfy a given memory budget. 
For example, descriptors can be allocated more space with light quantization but this comes at the cost of higher map compression, \ie, discarding more 3D points.

In this work, we propose a hybrid scene compression that efficiently quantizes descriptors and subsamples 3D points while maintaining the localization's score. 
Our approach focuses on optimizing descriptor memory efficiency with a quantization process that reaches extreme compression levels and still enables accurate descriptor matching and, consequently, precise localization.
By optimizing the memory allocated for descriptors, we can preserve more 3D points, thereby increasing the pool of 2D-3D matches available for localization.
A larger pool of accurate matches allows the proposed compression to preserve the localization performance, as long as the compression remains within reasonable limits.

The descriptors are compressed with a Differentiable Product Quantization (DPQ)~\cite{jegou2010product,chen2020differentiable} network, which achieves higher memory efficiency than traditional K-means or binary quantization~\cite{camposeco2019hybrid,jegou2010product,Yang_2022_CVPR}.
To compensate for the loss of information induced by the quantization process and to preserve matching accuracy, the map's quantized descriptors are `dequantized'  using a tiny-MLP decoder before matching, as demonstrated in~\cite{Yang_2022_CVPR}.
Compared to the binary quantization auto-encoder network in~\cite{Yang_2022_CVPR}, the proposed DPQ network is more memory efficient and is scene-specific to better fit the distribution of the scene's descriptors.

We train the DPQ in a standalone fashion with simple metric learning that not only recovers the dequantized descriptor but also preserves the relative distances between the original and dequantized descriptors, which is a key property for accurate matching. 
This is achieved by employing two triplet loss functions that operate on both the quantized and dequantized spaces.
Such training is simpler than previous DPQs~\cite{chen2020differentiable,chen2018learning} that are learned together with the downstream task, which is incompatible with the non-differentiable nature of structure-based localization~\cite{sarlin2019coarse}.
We also demonstrate how to combine the proposed descriptor compression with map compression to meet a given memory budget, and we show that localization is relatively insensitive to the proposed quantization. 
However, map compression is more limiting, as higher levels of compression reduce the pool of potential 2D-3D matches, no matter how accurate they are.
%

To summarize, this paper makes the following contribution:
i) we introduce a simple and standalone metric learning for Differentiable Product Quantization (DPQ) for scene compression that preserves matching properties of the descriptors and the final localization performance; 
ii)
the proposed hybrid method enables a better trade-off between memory complexity and localization than previous methods on several datasets;
iii) we quantitatively analyze the trade-offs between description and map compression and show how localization is more tolerant to description compression on outdoor and indoor datasets.

\begin{figure}[t!]
  \centering
  \includegraphics[width=\textwidth]{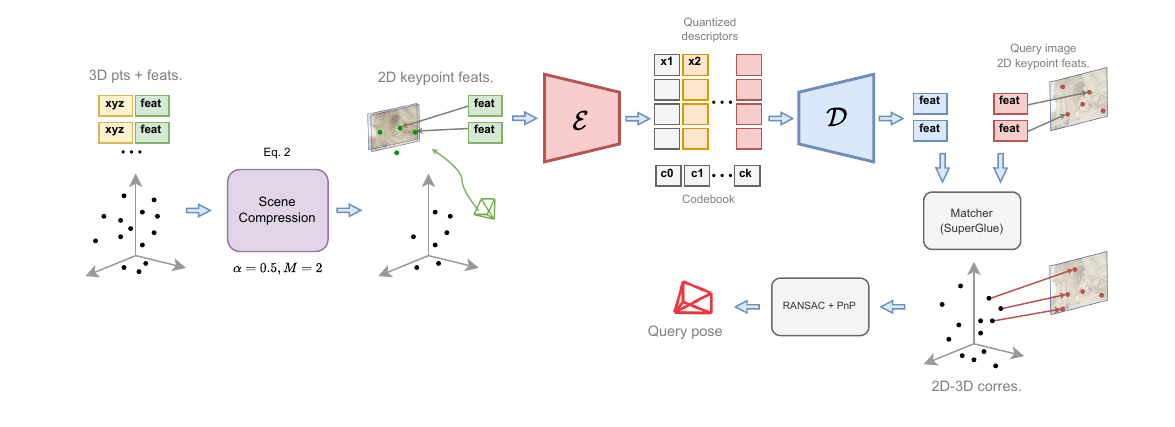}
  \caption{\textbf{Overview.} In this work we use differentiable Product Quantization to perform memory-efficient camera relocalization. More specifically, a set of local image descriptors extracted from an image is fed into an encoder $\mathcal{E}$ parameterized by the $M$ codebooks that are used to obtain a quantized representation of the input vectors. The quantized descriptors are then passed into the scene-specific differentiable decoder $\mathcal{D}$ that can recover the original descriptors that are susequently used in the localization pipeline. The encoder and decoder together represent a layer called D-PQED.}
  \label{fig:framework_overview}
\end{figure}

%% file: sections/related.tex
\section{Related Work} \label{sec:rel}

\boldparagraph{Visual Localization.}
In structure-based localization, the scene is represented by a map of 3D points and their descriptors.
To localize a query image, 2D-3D matches are drawn between the image and the map and fed to a PnP solver~\cite{1217599} that outputs the camera pose, usually within a robust estimation framework~\cite{fischler1981random, barath2020magsac++}. The map is built before the localization using SfM~\cite{schoenberger2016sfm} on a set of reference images: 2D image keypoints are extracted and described~\cite{lowe1999object, bay2006surf, calonder2011brief, rublee2011orb,yi2016lift, detone2018superpoint, dusmanu2019d2, revaud2019r2d2, liu2021dgd}, matched~\cite{lindenberger2023lightglue, shi2022clustergnn, sarlin2020superglue}, and triangulated into 3D points. The map comprises the 3D points tied to the 2D keypoints they get triangulated from and their descriptors. A typical method~\cite{sarlin2019coarse} to get 2D-3D matches efficiently between the map and the query image is to first derive 2D-2D matches between the query image and the most similar~\cite{arandjelovic2016netvlad,radenovic2018fine} reference images and lift the 2D reference points to their triangulated 3D points. Another one is Active Search (AS)~\cite{sattler2016efficient, sattler2012improving} that ranks the candidate points to match based on their visual word similarity. While structure-based localization sets the state-of-the-art~\cite{sarlin2019coarse}, their deployment is often limited by the memory requirements to store the 3D map.

A memory-efficient alternative is regression-based localization, \eg, absolute pose regression~\cite{kendall2015posenet,shavit2021learning,10236469,chen2023refinement} and scene coordinate regression (SCR)~\cite{brachmann2017dsac,li2020hierarchical,brachmann2021dsacstar,brachmann2023accelerated,wang2021continual,10044413,tang2023neumap,wang2024hscnet++}. It has a lower memory footprint since it implicitly represents the 3D scene with neural network.
However, the limited accuracy of pose regressor and the limited scalability of SCR still prevent their deployment in practical scenarios, hence the focus on the structure-based approaches.

\boldparagraph{3D Scene Compression.}
There are two main categories of 3D scene compression for visual localization: map compression, \ie, removing 3D points from the map, and descriptor compression.

Map compression~\cite{6909460, 7139575, li2010location, camposeco2019hybrid, Yang_2022_CVPR} strategically selects a subset of the map's 3D points to get a minimal representation of the scene that covers the same area as the original map. The process is framed as a K-covering problem that keeps a minimal set of points with high visibility~\cite{li2010location,camposeco2019hybrid} so that each reference image sees at least K 3D points, which enforces the spatial coverage of the scene. Selecting distinct points~\cite{6909460} or ensuring that the 3D points project uniformly onto the reference images~\cite{camposeco2019hybrid} are other relevant criteria. An alternative to the K-covering formulation is to frame the map compression problem using Mixed-Integer Quadratic Programming (QP)~\cite{mera2020efficient,soo20133d,dymczyk2015keep} that optimizes for coverage and distinctiveness and outputs a binary state for each point to be kept or discarded. To avoid defining the coverage and distinctiveness from heuristics~\cite{mera2020efficient}, SceneSqueezer~\cite{Yang_2022_CVPR} learn those heuristics from 3D features and solve the QP problem in a differentiable manner~\cite{agrawal2019differentiable}. This paper leverages the existing map compression of~\cite{mera2020efficient} and contributes to improving the descriptor compression.

In parallel to map compression, descriptor compression reduces the memory overhead of 
the descriptors while preserving their matching properties.
The main approaches involve dimensionality reduction~\cite{dong2023learning,valenzuela2012dimensionality,ke2004pca}, binarization~\cite{Yang_2022_CVPR}, K-means~\cite{7410600}, or hand-crafted product quantization (PQ)~\cite{lynen2015get,jegou2010product}. PQ usually reaches higher compression levels than other methods~\cite{Yang_2022_CVPR,jegou2008hamming}, which motivates its use in this work. The loss of information induced by the quantization impedes accurate descriptor matching so we train a tiny-MLP decoder to dequantize the descriptors before matching, as in~\cite{Yang_2022_CVPR}. This is achieved by learning a Differentiable version of PQ (DPQ) in an auto-encoder fashion.
Previous instances of DPQ~\cite{chen2020differentiable,chen2018learning} must be learned together with the downstream task, which is incompatible with the non-differentiable nature of structure-based localization~\cite{sarlin2019coarse}. Instead, we train DPQ with simple metric learning in a standalone optimization.

Hybrid methods use both map and descriptor compression to reach even tighter memory budgets although the descriptor matching accuracy drops~\cite{lynen2015get} as more information gets discarded. Camposeco~\etal~\cite{camposeco2019hybrid} address this issue by preserving a subset of the original descriptors for 2D-3D matching and using the quantized ones for geometric verification, but this comes at the cost of a higher memory footprint. Closest to this work, SceneSqueezer~\cite{Yang_2022_CVPR} compresses all descriptors but recovers the discarded information by training a network to quantize and dequantize the descriptors in an auto-encoder fashion. The proposed method also learns the descriptor's quantization and dequantization but differs from SceneSqueezer~\cite{Yang_2022_CVPR} in that it implements PQ~\cite{jegou2010product}, whose design enables tighter compression ratios~\cite{chen2018learning,chen2020differentiable}.
Also, this paper analyzes the interaction between the map and descriptor compression for a given memory budget and their effect on the localization, which is not investigated in~\cite{Yang_2022_CVPR}.

Extreme solutions to the descriptors' memory overhead discard all descriptors and run descriptor-free 2D-3D point matching~\cite{zhou2022geometry, wang2023dgc,campbell2020solving} but this comes at the cost of a large drop in localization performance. Our approach achieves a higher compression ratio and better localization. 

%% file: sections/preliminaries.tex
\section{Preliminaries}\label{sec:pre}
This section introduces the main concepts used in our hybrid map compression, namely product quantization, map compression, and low-rank adaptation.

\boldparagraph{Product Quantization (PQ).}
PQ~\cite{jegou2010product} splits a high-dimensional vector space into smaller, more manageable subspaces that are then quantized independently. A vector in the original high-dimensional space is transformed into the concatenation of the quantized codes from each subspace. Suppose we have a database $\{\bx\}$ of high-dimensional vectors $\bx \in \nR^D$, where $D$ is the dimensionality of the vector space. Each vector in the database is divided into $M$ sub-vectors, each of dimension $D' = D/M$, \ie $\bx=[\bx_1,\bx_2,...,\bx_M]$, where $\bx_m$ is the $m$-th sub-vector of $\bx$. For each subspace $m$, a codebook $\mathbf{C}_m \in \mathbb{R}^{K \times D'}$ of $K$ centroids is trained on $\{\bx_m\}$ using K-Means~\cite{bishop2006pattern}.
This results in $M$ codebooks $\mathbf{C} = [\mathbf{C}_1...\mathbf{C}_m...\mathbf{C}_M]$. Each sub-vector $\bx_m$ is then quantized independently into the index $q_m$ of the centroid closest to in $\bx_m$ the codebook $\mathbf{C}_m$:

\begin{equation}
\label{eq:final-loss}
q_m = \argmin_{\bc \in C_m} \| \bx_m - \bc \|
\end{equation}

\noindent The tuple of indices $(q_1,...q_m,...q_M)$ from all $M$ subspaces forms the quantized representation of the original vector. The original vector can be approximated by concatenating the centroids corresponding to these indices.

\boldparagraph{Map Compression ($\Gamma$).}
We adopt the map compression of~\cite{mera2020efficient}, also used in~\cite{Yang_2022_CVPR}, to select 3D points that are both distinctive and spatially far away from each other to provide uniform coverage of the scene. This is done by solving the following optimization problem where $v$ represents a discrete probability distribution over the 3D points that has non-zero values on the selected points:

\begin{equation}
\begin{aligned}
  \min_{v} \hspace{2mm} v^T K v - \tau d^T v \\
  \textrm{s.t.} \sum_{i}^{m} v_i = 1, \ \ \  0 \leq v_i \leq \frac{1}{\alpha m};
\end{aligned}
\label{eq:map_compr}
\end{equation}

\noindent $\alpha \in  (0,1]$ is the compression ratio with lower $\alpha$ resulting in higher map compression.
$K$ is the distance matrix storing the Euclidean distance between all pairs of 3D points, and $d$ is a vector that defines the distinctiveness of a 3D point as the fraction of total images (cameras) observing that 3D point.

\boldparagraph{Low-Rank Adaptation.} LoRA~\cite{hu2021lora} fine-tunes large models without the need to retrain the entire model, which can be computationally expensive and time-consuming. Instead of updating all the weights of a neural network, LoRA adapts only a subset of specific layers (\eg, linear layers) and introduces low-rank matrices that capture task-specific adaptations. Specifically, let us consider a weight matrix $W\in\nR^{d_\text{in}\times d_\text{out}}$ where $d_\text{in}$ and $d_\text{out}$ represent the input and output dimension, respectively. Instead of updating $W$ directly during fine-tuning, we keep $W$ fixed and introduce two matrices, $B\in\nR^{d_\text{in}\times r}$ and $A\in\nR^{r \times d_\text{out}}$, where $r$ is the rank which is much smaller than $d_\text{in}$ and $d_\text{out}$. The effective adaptation of the weight matrix $W$ is then computed as $W^{\prime}=W + BA$. During training, only low-rank matrices $A$ and $B$ are updated, and $W$ remains unchanged. 

%% file: sections/method.tex
\section{Method}\label{sec:method}

We propose a hybrid scene compression method to enable high-performance structure-based visual localization under memory constraints. After formalizing the problem, we introduce the quantization process run by an auto-encoder model that quantizes and dequantizes descriptors in an end-to-end manner in~\secref{ssec:model}.
Next, we describe the loss functions and the optimization process in~\secref{ssec:optimization}.
An overview of the proposed approach is provided in~\cref{fig:framework_overview}.

\boldparagraph{Problem Formulation}
Given a scene represented by a set of 3D points, their associated 2D image keypoints, and their local image descriptors, we generate a compressed representation of the scene suitable for localization
with a limited memory budget. In order to perform camera relocalization, we first build a map of the scene using SfM~\cite{schoenberger2016sfm}. Given a set of database images, 2D keypoints are extracted, matched, and triangulated into 3D points.

\subsection{D-PQED Layer}\label{ssec:model}
We develop a Differentiable Product Quantization with an Encoder-Decoder architecture (D-PQED) to quantize-dequantize image descriptors. A set of local image descriptors $\{\bx\}$ associated with 2D keypoints extracted from an image $I$ is quantized by the encoder $\be_\gamma$ and then passed to the decoder $\bd_\theta$.

\boldparagraph{Encoder.} The encoder $\be_\gamma$ is a differentiable PQ layer with weights $\gamma$ and parameterized by the $M$ codebooks $\bC$  = [$\bC_1,...\bC_m,...\bC_M$].
Each codebook holds $K$ centroids $\mathbf{c}_{mi} \in \bC_m$ of dimension $D'$ with $i \in \{0,K\}$, referred to as hard-quantized vector. The codebooks are initialized using standard PQ~\cite{jegou2010product} on the set of local descriptors $\{\bx\}$ (\cf~\secref{sec:pre}). For each vector $\bx$, PQ assigns the $m$-th subvector $\bx_m$ to the index $q_m \in \{0, K\}$ of the closest centroid in $\bC_m$(\eqnref{eq:final-loss}). This hard-assignment can be seen as a one-hot vector, $\mathbf{h}_m \in \{0,1\}^K$, where $\mathbf{h}_m(q_m)$ is 1 and 0 elsewhere. The quantized vector $\bq$ can be dequantized, \ie, the original vector $\bx$ can be approximated by a linear combination, noted $\Tilde{\bx}_m$, of the assignments $\mathbf{h}_m$ and the centroids $\mathbf{c}_{mi} \in \bC_m$. However, such approximations \{$\Tilde{\bx}$\} do not fully recover the original descriptors, which cause a severe drop in matching and localization performance, as shown in the results. Instead, we learn both quantization and dequantization to derive better approximations. To do so, the codebooks $\bC$ are optimized directly on \{$\bx$\} to obtain robust hard-quantized descriptors that minimize the error between the original and approximated descriptors. Since PQ involves a hard-assignment \eqnref{eq:final-loss} that is non-differentiable, we approximate the non-differentiable $\texttt{argmin}$ using a soft assignment during training and retain the hard assignment during inference. The soft assignment, noted $\ba_m$, is a soft-max operation on pairwise distances between sub-vector $\bx_m$ and the centroids, $\bc_{mi} \in \bC_m$. The soft-quantized vector, \ie the soft approximation of $\bx$, is obtained by a linear combination of $\ba_m$ and $\mathbf{c}_{mi}$.

We use the straight-through estimator~\cite{bengio2013estimating,yin2018understanding} to approximate the hard-quantization gradient: it adds the residual vector between hard and soft-quantized vector with a stop-gradient operation to the soft-quantized vector. This allows the backward pass to treat the quantization as an identity operation and ensure the gradients flow back through the network.

\input{algos/diff_pq_layer}

\boldparagraph{Decoder $\bd_\theta$.}
As mentioned previously, we learn the dequantization to reduce the loss of information induced by quantization, which is critical for localization. Inspired by the performance of scene-specific neural implicit models~\cite{mildenhall2021nerf} and scene-coordinate regression models~\cite{brachmann2021dsacstar,wang2024hscnet++}, we train a scene-specific dequantization decoder. The decoder $\bd_\theta$ is trained to dequantize the quantized descriptors from $\be_\gamma$ and ideally satisfies $\bd_\theta(\be_\gamma(\bx))=\hat{\bx}$ for all descriptors $\bx$ extracted from an image, \ie, minimize the error $\| \bx - \hat{\bx}\|$. 

We parametrize $\bd_\theta$ with a two-layer MLP and weights $\theta$. During inference, only the non-zero index ($q_m$) in $\bh_m$ is stored per subspace, resulting in a memory of $M \times \mathrm{log_2} (K)$ bytes per vector. This is much more memory-efficient than differentiable binary quantization~\cite{jegou2008hamming,Yang_2022_CVPR}, which requires $D / 8$ bytes per vector. 

\subsection{Optimization}\label{ssec:optimization}
One of the criteria to optimize the network is to minimize the distance between the original descriptors and their approximation, \eg, using the L2 loss~\cite{Yang_2022_CVPR}. To further preserve the descriptor matching performance, and hence localization, we enforce a second constraint for the network to maintain the ranking between the original and approximated descriptors. This is implemented with a triplet margin loss~\cite{schroff2015facenet,schultz2003learning,wang2014learning,weinberger2009distance} in both the original (raw) and dequantized space, encouraging the distance between negative samples to be larger than the distance between positive samples by a margin $m$. For a descriptor $\bx$, we define the positive sample as the unquantized descriptor $\bd_\theta(\be_\gamma(\bx))$ so the positive distance is simply the approximation error. Following~\cite{mishchuk2017working}, we use in-batch hard negative mining to select the closest negative, \ie, the unquantized descriptor $\bd_\theta(\be_\gamma(\bx))$ closest to $\bx$ but that is not related to $\bx$. These two descriptors should be far from each other in both the original space and the dequantized space, which is implemented with two loss terms $\mathcal{L_\text{raw}}$ and $\mathcal{L_\text{d}}$.

The first term $\mathcal{L_\text{raw}}$ expresses the negative distance $\text{neg}_{\text{raw}}$ as the distance between the dequantized descriptor and the hardest original descriptor. In the second term, $\mathcal{L_\text{d}}$, it is the distance $\text{neg}_{\text{d}}$ between the dequantized descriptor and the hardest dequantized descriptor. More formally, 

\begin{equation}
    \mathcal{L_\text{raw}} = \frac{1}{|\mathcal{X}|}\sum_{\bx \in \mathcal{X}}\text{max}(m + \text{pos}(\bx) - \text{neg}_{\text{raw}}(\bx), 0)
\end{equation}

\begin{equation}
    \mathcal{L_\text{d}} = \frac{1}{|\mathcal{X}|}\sum_{\bx \in \mathcal{X}}\text{max}(m + \text{pos}(\bx) - \text{neg}_{\text{d}}(\bx), 0)
\end{equation}

\noindent where $\text{pos}(\bx) = \|\bx - \bd_\theta(\be_\gamma(\bx)) \|$ is the positive distance. The negative distance in the original space and dequantized space are defined as follows:

\begin{equation}
    \text{neg}_{\text{raw}}(\bx) = \min_{\bx^{\prime} \in \mathcal{X}, \bx^{\prime} \ne \bx} \|\bx^{\prime} - \bd_\theta(\be_\gamma(\bx)) \|
\end{equation}

\begin{equation}
    \text{neg}_{\text{d}}(\bx) = \min_{\bx^{\prime} \in \mathcal{X}, \bx^{\prime} \ne \bx} \|\bd_\theta(\be_\gamma(\bx)) - \bd_\theta(\be_\gamma(\bx^{\prime})) \|
\end{equation}

The final loss is a weighted sum of the triplet ranking loss in the original and unquantized space and defined as: 
\begin{equation}
    \mathcal{L}=\mathcal{L}_\text{raw} + \lambda_1\mathcal{L}_\text{d}
\end{equation}

\input{tbl/aachen_bline_m}

\input{tbl/pq_alpha_dpqed}

\boldparagraph{Implementation details.}
The initial 3D map, consisting of 3D points and local descriptors, is generated with Structure-from-Motion from SuperPoint~\cite{detone2018superpoint} keypoints $\bx \in \nR^{256}$ matched with SuperGlue~\cite{sarlin2020superglue} and triangulated with COLMAP~\cite{sarlin2019coarse,schoenberger2016sfm}.
Given the 3D map, we first quantize the descriptors at a desired PQ level ($M$) then the decoder is trained to de-quantize them.
We train a simple 2-layer MLP with a hidden dimension of 256 that is implemented in PyTorch~\cite{paszke2019pytorch} with the Adam optimizer~\cite{KingBa15}, batch size of 1000 and a learning rate of 0.001. The training is done with a triplet margin of 0.9 for 30 epochs. During training, the softmax assignment at the output of PQ layers is scaled by $\tau$ = 0.05. Given a memory budget $B$, only a subset of the quantized descriptors are stored in memory by using map compression $\Gamma_\alpha$~\cite{mera2020efficient}, \ie, by removing 3D points. The compression ratio is obtained by $\alpha = B/B_M$ where $B_M = M \times \mathrm{log_2} (K)$ is the memory occupied by descriptors at PQ level $M$, noted PQ$M$. In the paper, we provide results for $M$ = 2,4,32 and $K$ = 256. For more implementation details, please refer to the supplementary material.

%% file: algos/diff_pq_layer.tex
\noindent\begin{figure}[t!]
\centering
\resizebox{0.8\textwidth}{!}{
\begin{minipage}{\textwidth}
\scriptsize
\begin{algorithm}[H]
\caption{Forward Pass of the encoder $\be_\gamma$ of the proposed D-PQED Layer with the Straight-Through trick}
\begin{algorithmic}[1]
\Require Input vector $\mathbf{x} \in \mathbb{R}^D$, number of subspaces $M$, number of centroids $K$ per subspace, initialized centroids $\mathbf{C}_m$ for each subspace $r$, where $\mathbf{C}_m \in \mathbb{R}^{K \times D'}$ and $D' = D / M$
\Ensure Quantized representation of $\mathbf{x}$ with gradient support via straight-through estimator


\For{$m = 1$ to $M$}
    \State $\mathbf{x}_m \gets$ Extract $m$-th subspace from $\mathbf{x}$ \Comment{$\mathbf{x}_m \in \mathbb{R}^{D'}$}
   \For{$i = 1$ to $K$}
        \State Store in $\mathbf{d}_{m} \gets \| \mathbf{x}_m - \mathbf{c}_{mi} \|_2$ \Comment{$\mathbf{c}_{mi}$ is $i$-th centroid in $m$-th subspace}
    \EndFor
    \State $\mathbf{a}_m \gets$ softmax($-\mathbf{d}_{m}$) \Comment{Soft assignment}
    \State $\mathbf{h}_m \gets$ argmin($\mathbf{d}_{m}$) \Comment{Hard assignment}
    \State $\dot{\mathbf{x}}_m \gets \sum_{i=1}^{K} a_{mi} \mathbf{c}_{mi}$ \Comment{Soft quantized vector}
    \State $\tilde{\mathbf{x}}_m \gets \sum_{i=1}^{K} h_{mi} \mathbf{c}_{mi}$ \Comment{Hard quantized vector}
    \State $\dot{\mathbf{x}}_m \gets \dot{\mathbf{x}}_m + \text{stop\_gradient}(\tilde{\mathbf{x}}_m - \dot{\mathbf{x}}_m)$ \Comment{Straight-through}
\EndFor

\State $\dot{\mathbf{x}} \gets$ Concatenate($\dot{\mathbf{x}}_1, \dot{\mathbf{x}}_2, \ldots, \dot{\mathbf{x}}_M$) \Comment{Quantized representation of $\mathbf{x}$}

\Return $\dot{\mathbf{x}}$
\end{algorithmic}\label{alg:diff-pq-layer}
\end{algorithm}
\end{minipage}
}
\vspace{-0.5cm}
\end{figure}


%% file: tbl/aachen_bline_m.tex
\begin{table}[t!]
\caption{\textbf{Evaluation of Different Compression techniques.} We evaluate Product Quantization~\cite{jegou2010product} PQ$M$ with $M$ the number of quantization levels), and map compression ($\Gamma_\alpha$)~\cite{mera2020efficient} on two localization datasets, Aachen Day-Night~\cite{Sattler2012BMVC,sattler2018benchmarking} and ETH Microsoft~\cite{eth_ms_visloc_2021}. 
We use the $/$ symbol to report the memory consumption, MB and the corresponding compression ratios $\alpha$ used for the Aachen and ETH datasets, respectively.}
\label{tab:aachen-compression-methods}
\centering\scriptsize
\setlength{\tabcolsep}{3.4pt}

\begin{tabular}{lrcccccc|cccccc}  
\toprule
{} & {} & \multicolumn{6}{c|}{Aachen Day-Night~\cite{Sattler2012BMVC,sattler2018benchmarking}} & \multicolumn{6}{c}{ETH Microsoft~\cite{eth_ms_visloc_2021}} \\
 & & \multicolumn{6}{c|}{$(0.25m,2^\circ)/(0.5m,5^\circ)/(1.0m,10^\circ)$} & \multicolumn{6}{c}{$(0.10m,1^\circ)/(0.25m,2^\circ)/(1.0m,5^\circ)$} \\
 & MB & \multicolumn{3}{c}{day-time} & \multicolumn{3}{c|}{night-time } & \multicolumn{3}{c}{single} & \multicolumn{3}{c}{multi-rig} \\
\midrule
HLoc~\cite{sarlin2019coarse} & 6330/4463 & 88.5 & 95.5 & 98.7 & 85.7 & 92.9 & 100.0 & 46.3 & 55.0 & 62.3 & 74.7 & 75.7 & 76.0 \\
HLoc-avg & 618/368 & 88.5 & 95.8 & 98.8 & 84.7 & 93.9 & 100.0 & 46.0 & 53.7 & 64.0 & 75.3 & 79.0 & 80.0 \\
PQ32 & 38/23 & 88.6 & 95.9 & 98.7 &	83.7 & 93.9 & 100.0 & 43.7 & 51.0 & 57.7 & 75.3 & 79.3 & 80.7 \\
$\Gamma_{.063/.064}$ & 38/23 & 83.9 & 92.5 & 96.5 &	77.6 & 86.7 & 98.0 & 25.7 & 35.0 & 43.3 & 57.3 & 65.3 & 70.0 \\
PQ4 & 4/2.87 & 81.2 & 88.6 & 92.8 &	45.9 & 53.1 & 62.2 & 16.3 & 18.3 & 20.7 & 67.0 & 69.7 & 71.3 \\
$\Gamma_{.006/.008}$ & 4/2.87 & 50.5 & 61.2 & 70.5	 &31.6 & 43.9 & 57.1 & 8.0 & 11.7 & 16.7 & 29.0 & 36.3 & 43.7 \\
PQ2 & 2/1.43 & 69.9 & 76.9 & 82.0 &	30.6 & 40.8 & 45.9 & 8.7 & 11.0 & 12.3 & 58.3 & 63.3 & 65.3 \\
$\Gamma_{.003}$ & 2/1.43 & 36.5 & 45.6 & 55.2 &	12.2 & 20.4 & 40.8 & 4.0 & 7.7 & 9.0 & 21.0 & 24.7 & 31.7\\
\bottomrule
\end{tabular}
\end{table}

%% file: tbl/pq_alpha_dpqed.tex
\begin{table}[t!]
\caption{\textbf{Camera Relocalization: Extreme Memory Budget.} The proposed D-PQED layer provides significantly better localization performance for challenging scenes compared to traditional 3D scene compression techniques.}\label{tab:pq-vs-compression-localization}
\centering\scriptsize
\setlength{\tabcolsep}{2.9pt}

\begin{tabular}{lrcccccc|cccccc}  
\toprule
{} & {} & \multicolumn{6}{c|}{Aachen Day-Night~\cite{Sattler2012BMVC,sattler2018benchmarking}} & \multicolumn{6}{c}{ETH Microsoft~\cite{eth_ms_visloc_2021}} \\
 & & \multicolumn{6}{c|}{$(0.25m,2^\circ)/(0.5m,5^\circ)/(1.0m,10^\circ)$} & \multicolumn{6}{c}{$(0.10m,1^\circ)/(0.25m,2^\circ)/(1.0m,5^\circ)$} \\
 & MB & \multicolumn{3}{c}{day-time} & \multicolumn{3}{c|}{night-time } & \multicolumn{3}{c}{single} & \multicolumn{3}{c}{multi-rig} \\
\midrule
PQ32+$\Gamma_{.025/.043}$ & 1 & 71.5 & 81.2 & 85.6 & 53.1 & 69.4 & 79.6 & 19.0 & 24.7 & 28.3 & 55.7 & 62.0 & 67.0 \\
+D-PQED & 1 & 75.6 & 84.8 & 88.7 & 62.2 & 73.5 & 85.7 & 19.7 & 29.3 & 35.0 & 52.3 & 63.3 & 68.0 \\
PQ4+$\Gamma_{.25/.348}$ & 1 & 76.3 & 83.1 & 88.6 & 36.7 & 45.9 & 49.0 & 10.0 & 13.0 & 15.7 & 55.3 & 59.7 & 64.3 \\
+D-PQED & 1 & \textbf{86.0} & \textbf{93.2} & \textbf{96.8} & \textbf{77.6} & \textbf{87.8} & \textbf{96.9} & \textbf{30.0} & \textbf{38.3} & \textbf{46.3} & \textbf{67.0} & \textbf{72.3} & \textbf{74.3} \\
PQ2+$\Gamma_{.5/.699}$  &	1 &	65.0 & 73.2 & 78.0 &	26.5 & 32.7 & 33.7 & 5.0 & 7.7 & 9.0 & 54.3 & 60.3 & 64.0 \\
+D-PQED  &	1 &	84.7 & 92.1 & 95.4 & 66.3 & 76.5 & 86.7 & 22.3 & 31.0 & 39.0 & 64.3 & 69.7 & 73.0 \\
\midrule
PQ32+$\Gamma_{.013/.022}$ & 0.5 & 60.4 & 70.5 & 78.6 & 43.9 & 61.2 & 68.4 & 12.3 & 17.7 & 20.7 & 41.3 & 51.7 & 57.3 \\
+D-PQED & 0.5 & 63.0 & 74.2 & 82.0 & 53.1 & 66.3 & 77.6 & 14.0 & 20.3 & 25.0 & 46.0 & 51.7 & 57.0 \\
PQ4+$\Gamma_{.125/.174}$ & 0.5 & 71.7 & 79.0 & 84.3 & 32.7 & 43.9 & 48.0 & 8.0 & 10.0 & 13.3 & 51.0 & 56.3 & 60.3 \\
+D-PQED & 0.5 & \textbf{84.5} & \textbf{92.1} & \textbf{96.1} & \textbf{68.4} & \textbf{81.6} & \textbf{92.9} & \textbf{25.0} & \textbf{35.3} & \textbf{41.7} & \textbf{60.0} & \textbf{68.3} & \textbf{71.3} \\
PQ2+$\Gamma_{.25/.35}$  & 0.5 & 61.5 & 69.2 & 74.2 & 22.4 & 25.5 & 30.6 & 5.3 & 7.3 & 9.0 &	51.0 & 54.0 & 57.0 \\
+D-PQED & 0.5 & 82.5 & 91.0 & 94.5 &	59.2 & 70.4 & 82.7 & 18.3 & 25.7 & 30.7 &	62.0 & 66.3 & 69.0\\
\bottomrule
\end{tabular}
\end{table}

%% file: sections/experiments.tex
\section{Experiments}\label{sec:exp}
We evaluate the proposed quantization strategy on camera relocalization and compare against state-of-the-art scene neural mapping and compression techniques as well as recent feature-based localization methods. We also provide an extensive ablation study on different design choices of our approach.

\boldparagraph{Datasets.} 
We evaluate the proposed approach based on the performance of indoor and outdoor camera relocalization on 4 standard benchmarks:
a) Aachen Day-Night~\cite{Sattler2012BMVC,sattler2018benchmarking}: a real-world city-scale outdoor dataset with day-time database and night-time query images, respectively; b) Cambridge landmarks~\cite{kendall2015posenet}: a collection of 5 medium scale outdoor scenes; c) 7Scenes dataset~\cite{shotton2013scene}: a real-world indoor dataset with 7 small disjoint scenes; d) ETH-Microsoft Dataset~\cite{eth_ms_visloc_2021}: it spans day and night, large indoor and outdoor environments, various sensor setups, and both single and multi-rig queries. The database images are used to build the 3D map used at test time for localization by structure-based methods~\cite{sarlin2019coarse}, and also used for training networks.

\boldparagraph{Metrics.} For each query image, the errors $e_R$ and $e_t$ of the estimated rotation and translation, $\hat{\bR}$ and $\hat{\bt}$, are computed as $e_R$ = $\text{cos}^{-1}(0.5 \times \text{Tr}(\hat{\bR}^\text{T}\bR)-1)$ and $e_t$ = $||\hat{\bR}^\text{T} \hat{\bt} - \bR^T\bt ||_2^2$, where $\text{Tr}$ is the trace operator; $\bR$ is a rotation matrix, and $\bt$ is translation.
Given the rotation and translation errors, two metrics are reported~\cite{kendall2015posenet}: a) accuracy and b) median rotation/translation errors.
The accuracy is the percentage of queries whose translation and rotation errors are below a threshold $(t/r)$.
For Aachen~\cite{sattler2018benchmarking,Sattler2012BMVC}, the accuracy is reported over three thresholds $(0.25/2), (0.5/5), (5,10) \text{m}/^o$. For Cambridge landmarks~\cite{kendall2015posenet}, we report the median of the translation and rotation errors.
For 7Scenes~\cite{shotton2013scene}, we provide both median pose errors and accuracy with a threshold of $(0.05/5)$ $\text{m}/^o$ .

\boldparagraph{Baselines.} We compare against: a) state-of-the-art regression-based methods: PixLoc~\cite{sarlin2021back}, GLACE~\cite{wang2024glace},  and NeuMap~\cite{tang2023neumap}; b) descriptor-free methods: DGC-GNN~\cite{wang2023dgc}, Go-Match~\cite{zhou2022geometry}; c) compression methods: Scene-Squeezer~\cite{Yang_2022_CVPR}, QP+R.Sift~\cite{mera2020efficient}, Cascaded~\cite{cheng2019cascaded}; d) structure-based methods: HLoc~\cite{sarlin2019coarse} and Active-Search (AS)~\cite{sattler2016efficient}. 
We also propose 3 competitive baselines:
e) HLoc-avg: for each 3D point,
we discard the local descriptors of the 2D keypoints in the 3D point's track and instead keep only their average~\cite{li2010location}. This also occurs in methods that perform 2D-3D matching, \eg, Active-Search~\cite{sattler2016efficient}. f) the map compression of~\cite{mera2020efficient} $\Gamma_\alpha$ with $\alpha$ the compression ratio presented in Eq.~\ref{eq:map_compr}. We consider this method a baseline since the official implementation of~\cite{mera2020efficient} is not publicly available; g) off-the-shelf product-quantization~\cite{jegou2010product} with various compression levels.

\input{tbl/main_results}

\subsection{Results}

\noindent\textbf{PQ vs. map compression vs. D-PQED.}
We first evaluate different scene compression methods on the Aachen Day-Night~\cite{sattler2018benchmarking,Sattler2012BMVC} and ETH Microsoft~\cite{eth_ms_visloc_2021} localization datasets. Localization results and local descriptor memory are reported in~\tabref{tab:aachen-compression-methods}.
Compared to HLoc~\cite{sarlin2019coarse}, HLoc-avg reduces the memory requirement by an order of magnitude while retaining localization accuracy.
This is a key observation as it makes the memory comparable to AS~\cite{sattler2016efficient} (\cf~\tabref{tab:main_results}) while having much better performance.

Baselines based on product quantization, PQ$M$~\cite{jegou2010product}, produce surprisingly accurate results, especially in daytime queries.
We observe that for PQ32, the localization remains competitive and requires only 38MB of local descriptor memory.
For the same memory budget, map compression~\cite{mera2020efficient} $\Gamma_\alpha$ performs worse than PQ-based descriptor quantization.

We experiment with various levels of quantization and observe that as we increase the quantization by lowering $M$, the memory requirement decreases, but so does the localization performance.
As shown in~\tabref{tab:pq-vs-compression-localization}, combining descriptor quantization and map compression provides a good balance to reach a given memory budget and keep descriptors at a lower quantization level ($M \uparrow$). The combination of map compression $\Gamma_\alpha$~\cite{mera2020efficient} and PQ32 at a memory budget of 1MB outperforms stand-alone PQ2 (2MB).
Results are comparable only for multi-rig queries in the ETH Microsoft dataset. Similar benefits are observed for PQ4+$\Gamma_\alpha$(1MB) compared to PQ2 (2MB).

Our proposed method (PQ$M$+$\Gamma_\alpha$+D-PQED) further dequantizes the descriptors before matching and shows consistent and significant improvements across the majority of the accuracy thresholds for all PQ levels. Interestingly, a maximum in localization performance is observed for (PQ4+$\Gamma_\alpha$+D-PQED) under both 0.5 and 1MB memory budgets.
It is a maximum because the other variants (PQ32+$\Gamma_\alpha$+D-PQED) have lower quantization and higher compression ($M \uparrow, \alpha \downarrow$), while (PQ2+$\Gamma_\alpha$+D-PQED) has higher quantization and lower compression ($M \downarrow, \alpha \uparrow$). Our results suggest that for the best localization performance, prioritizing the storage of distinctive descriptors ($M \uparrow$) or a large number of 3D points ($\alpha \uparrow$) is not optimal. The best results can be obtained by tuning $\alpha, M$ to retain a sufficient number of 3D points that are distinctive enough for robust keypoint matching. This aligns with the trade-off between descriptor quantization and map compression: keeping distinctive descriptors comes at the cost of retaining fewer 3D points for matching ($M \uparrow, \alpha \downarrow$), whereas keeping more 3D points implies higher quantization of descriptors ($M \downarrow, \alpha \uparrow$), reducing the descriptor distinctiveness.

\noindent \textbf{State-of-the-art compression methods.} We now compare the proposed approach with existing methods in~\tabref{tab:main_results}. Different methods store different forms of database information so we report the total scene-specific memory and ignore scene-agnostic components of the respective methods. For scene-coordinate-based methods GLACE~\cite{wang2024glace} and NeuMap~\cite{tang2023neumap}, we include the dataset-specific model size in the memory.
For PixLoc~\cite{sarlin2021back}, which stores database images, we provide an approximate memory value. For descriptor-free methods Go-Match~\cite{zhou2022geometry} and DGC-GNN~\cite{wang2023dgc}, we report the 3D point cloud memory, including color information per 3D point for DGC-GNN. For SIFT-based structured localization methods such as AS~\cite{sattler2016efficient}, Cascaded~\cite{cheng2019cascaded}, QP + R.Sift~\cite{mera2020efficient} we report the memory from local descriptors and 3D points. For our approach, Ours (PQ4+$\Gamma_{0.25}$+D-PQED), HLoc~\cite{sarlin2019coarse}, and Scene-Squeezer~\cite{Yang_2022_CVPR}, we report the total memory from local descriptors, 3D points, and global descriptors.
Additionally, for Ours* (PQ4+$\Gamma_{0.25}$+D-PQED), we product quantize global descriptors with $M$ = 256. For both versions of our method, we include all (local/global) PQ codebooks in the total memory. Results for Aachen~\cite{Sattler2012BMVC,sattler2018benchmarking} and Cambridge~\cite{kendall2015posenet} show that after PQ of global descriptors, Ours* retains the same performance while reducing the memory requirements manyfold. As nearest-neighbors are already provided for the 7Scenes~\cite{shotton2013scene} dataset in the HLoc~\cite{sarlin2019coarse} evaluation pipeline, we do not include global descriptor memory in HLoc and D-PQED. However, if we did, the total memory for Ours would be approximately 200MB and 15MB fr Ours*. Based on the results on Aachen~\cite{Sattler2012BMVC,sattler2018benchmarking} and Cambridge~\cite{kendall2015posenet}, we would assume similar performance between Ours and Ours* on 7Scenes~\cite{shotton2013scene} as well.

Results show that our methods achieve the best memory-performance trade-off, with the second-best results at the lowest memory budget on the challenging Aachen Day-Night dataset~\cite{Sattler2012BMVC,sattler2018benchmarking}.
Interestingly, Ours* outperforms dataset-specific model-based approaches such as GLACE~\cite{wang2024glace} and NeuMap~\cite{tang2023neumap} both in terms of memory and localization performance. Results on Cambridge~\cite{kendall2015posenet} and 7Scenes~\cite{shotton2013scene} follow a similar trend: D-PQED methods achieve the best or second-best translation results and closely follow the best results for rotation estimation. 
Detailed scene-wise results on Cambridge and 7Scenes are presented in the Supplementary material. Surprisingly, even the PQ~\cite{jegou2010product} baselines do not suffer any significant drop in performance, indicating the presence of redundant information.

We experiment with an extreme memory budget of $B$ = 0.3MB.
Due to tight memory constraints and given that the baseline PQ~\cite{jegou2010product} already performs on par with the full pipeline, we do not fine-tune our decoder. Instead, we finetune the decoder trained on the Aachen Day-Night~\cite{Sattler2012BMVC,sattler2018benchmarking} dataset and added and updated LoRA~\cite{hu2021lora} weights, resulting in an overhead of only 0.008MB per scene. It is important to note that this is done to be fair in terms of including network weights in memory calculation.

We observe that both PQ2 and PQ4 in combination with D-PQED perform well. Qualitative results showing pixel correspondences obtained by PQ4 and the proposed approach are provided in~\figref{fig:qualitative_pq64_dpqed64_aachen_cambridge}.
\input{fig/qualitative_results}
\input{tbl/ablations}

\subsection{Ablation Study}
In this section, we analyze each design choice and its contribution. We compare the combined triplet margin loss against vanilla L2 reconstruction loss used in Scene-Squeezer~\cite{Yang_2022_CVPR}. Results in~\tabref{subtab:hloc-vs-pq1} show that triplet loss provides a significant boost in performance (\cf row 3 and row 4). Differentiable PQ layers offer substantial performance gains compared to off-the-shelf PQ~\cite{jegou2010product}. The PQ layers also complement the decoder, providing consistent improvements. Finally, we analyze the case of asymmetric localization, where queries are represented in the original representation space, while database descriptors are represented in the dequantized space. We additionally quantize and dequantize the query descriptors using the proposed D-PQED layer resulting in a symmetric setup. The results show that this is slightly worse than the asymmetric setup. We hypothesize that the quantization induces a loss of information that cannot be completely recovered with the decoder.
Query-side quantization results in even more information loss.

%% file: tbl/main_results.tex
\begin{table}[t!]
\caption{\textbf{Camera Relocalization: Benchmark.} We compare localization performance with state-of-the-art methods on the Aachen Day-Night, 7Scenes, and Cambridge Landmarks datasets. Ours and Ours*, both with (PQ4 + $\Gamma_{0.25}$ + D-PQED), outperform other map compression methods in terms of localization accuracy and memory budget. The best and second best results are marked with \textbf{bold} and \underline{underline}. The map compression methods are highlighted in \textit{italic}. 
}

\scriptsize
\centering
\resizebox{\textwidth}{!}{
\setlength{\tabcolsep}{4pt}
\renewcommand*{\arraystretch}{1.2}
\begin{tabular}{lrc|rcc|rc}  
\toprule
		&  \multicolumn{2}{c}{Aachen Day/Night~\cite{Sattler2012BMVC,sattler2018benchmarking}} & \multicolumn{3}{c}{7Scenes~\cite{shotton2013scene}} & \multicolumn{2}{c}{Cambridge~\cite{kendall2015posenet}} \\
   &  MB & Acc.,\%  @ (0.25,2) $\uparrow$ & MB & $\text{t,m / r},^\circ$ $\downarrow$ & Acc.,\% $\uparrow$ & MB & $\text{t,m / r},^\circ$ $\downarrow$ \\

\midrule
    HLoc~\cite{sarlin2019coarse, sarlin2020superglue} & 6330 &	\textbf{89.6} /  \textbf{86.7} & 9687 & \textbf{0.03} / \textbf{0.93} & \underline{76.8} & 5440 &  \textbf{0.09} / \textbf{0.23} \\
    AS~\cite{sattler2016efficient} &750 & 85.3 / 39.8& >500 & \underline{0.04} / 1.18 & 68.7 &813 & \underline{0.11} / \underline{0.26}  \\
    \textit{Cascaded}~\cite{cheng2019cascaded} & 140 & 76.7 / 33.7 & - & -  & - & - & -  \\
    \textit{QP+R.Sift} ($\Gamma_\alpha$)~\cite{mera2020efficient} & 31 & 62.6 / 16.3 & - & - &-& \underline{7.01} & 0.93 / 1.39 \\
    GLACE~\cite{wang2024glace} & \underline{23} & 9.8 / - & \underline{28}& - & \textbf{81.4}& 52& 0.17 / 0.30 \\
    PixLoc~\cite{sarlin2021back} & >2000 & 64.3 / 51.0 & >1000 & \textbf{0.03} / \underline{0.98} & 75.7& >1000 & \underline{0.11} / 0.28\\
    NeuMap~\cite{tang2023neumap} & 1260 & 80.8 / 48.0 & 70 & \textbf{0.03} / 1.09 & -& 75 & 0.14 / 0.33 \\
    \textit{Scene-Squeezer}~\cite{Yang_2022_CVPR} & 31 & 75.5 / 50.0 & - & - & - & \bf{1.91} & 0.23 / 0.42 \\
    Go-Match~\cite{zhou2022geometry} & -& -& 302& 0.22 / 5.78 &-& 48& 1.73 / 5.97\\
    DGC-GNN~\cite{wang2023dgc} & -& -& 355& 0.15 / 4.47 &-& 69& 0.54 / 2.23\\
    \midrule
    Ours &  40.75 & \underline{86.0} / \underline{77.6} & \textbf{1.64}* & \textbf{0.03} / 1.08 & 73.0 & 32.40 & \underline{0.11} / 0.28 \\
    Ours* & \textbf{6.96}  & 85.7 / \underline{77.6} & - & - &- & 8.85 &  \underline{0.11} / 0.27 \\

\bottomrule

\end{tabular}
}
\label{tab:main_results}
\end{table}

%% file: fig/qualitative_results.tex
\begin{figure}[h!]
\centering
\begin{subfigure}{.49\textwidth}
  \centering
  \includegraphics[width=\linewidth]{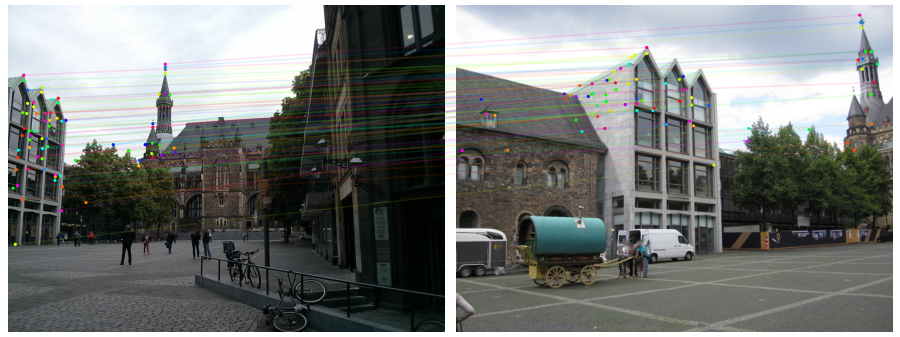}
  \includegraphics[width=\linewidth]{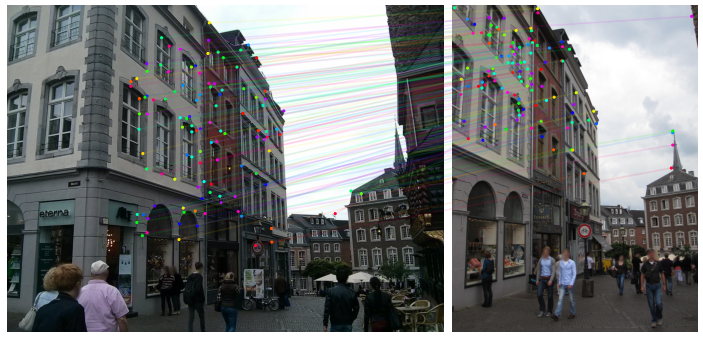}
  \includegraphics[width=\linewidth]{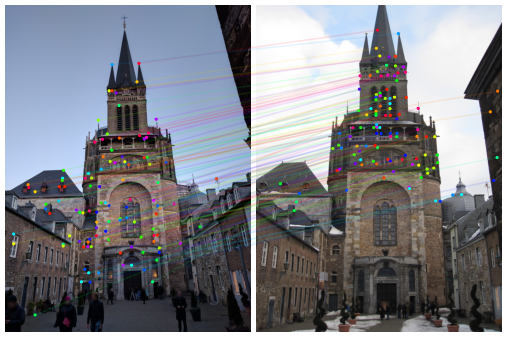}
  \includegraphics[width=\linewidth]{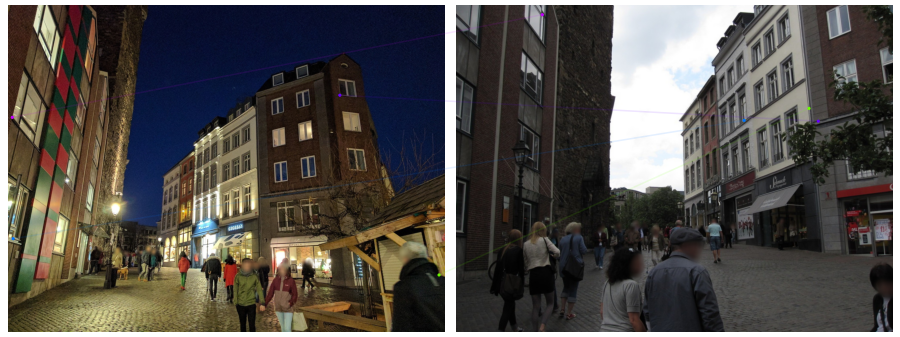}
  \includegraphics[width=\linewidth]{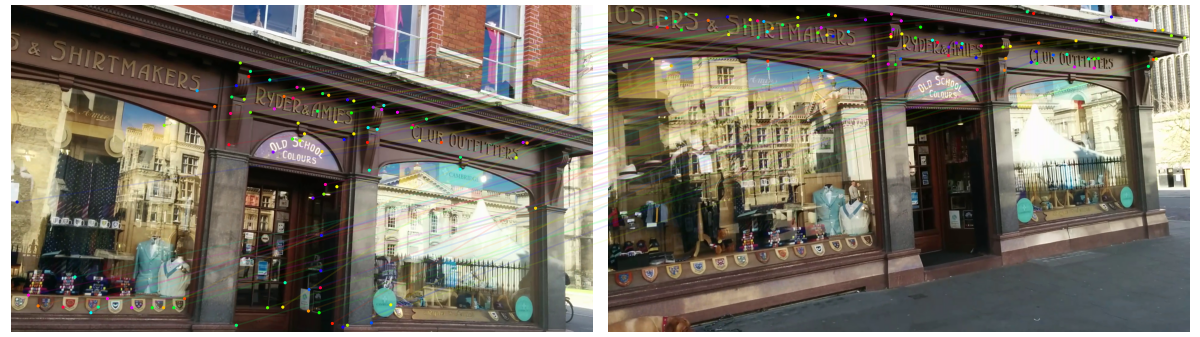}
  \includegraphics[width=\linewidth]{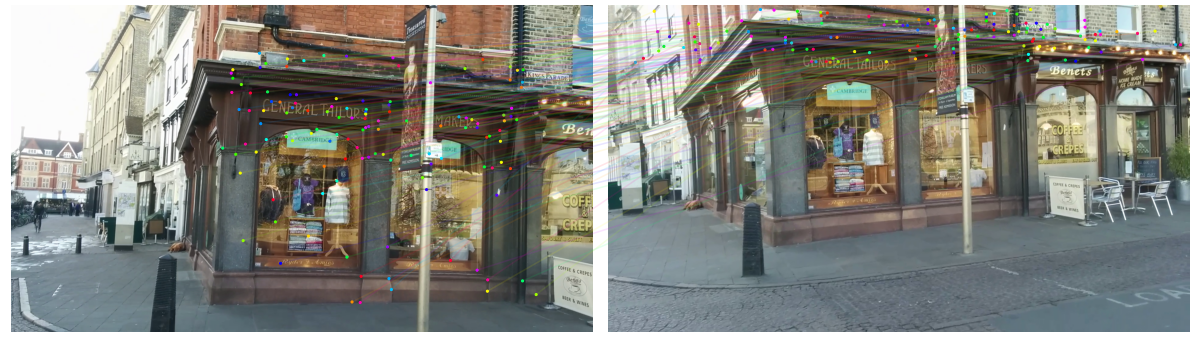}
  \caption{PQ4}
\end{subfigure}%
\hfill
\begin{subfigure}{.49\textwidth}
  \centering
  \includegraphics[width=\linewidth]{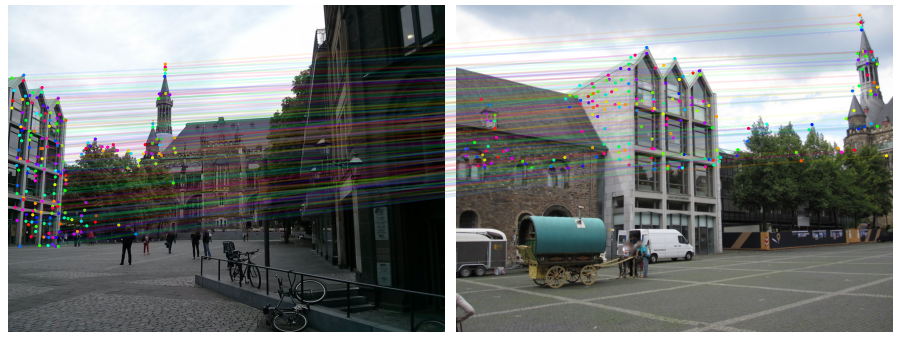}
  \includegraphics[width=\linewidth]{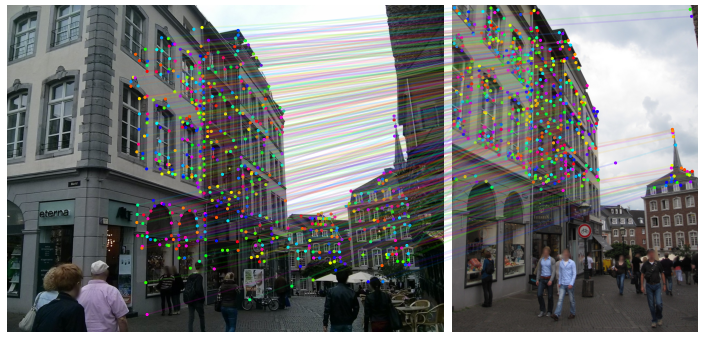}
  \includegraphics[width=\linewidth]{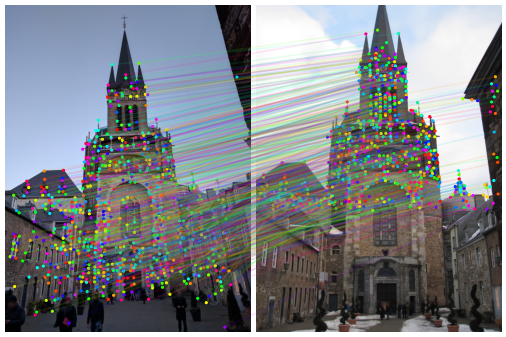}
  \includegraphics[width=\linewidth]{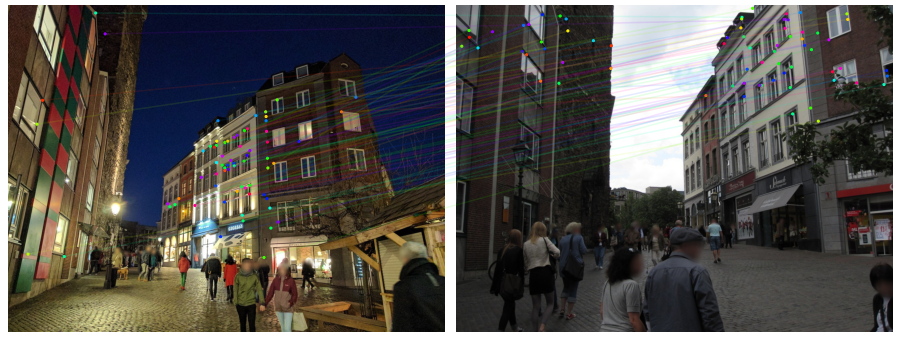}
  \includegraphics[width=\linewidth]{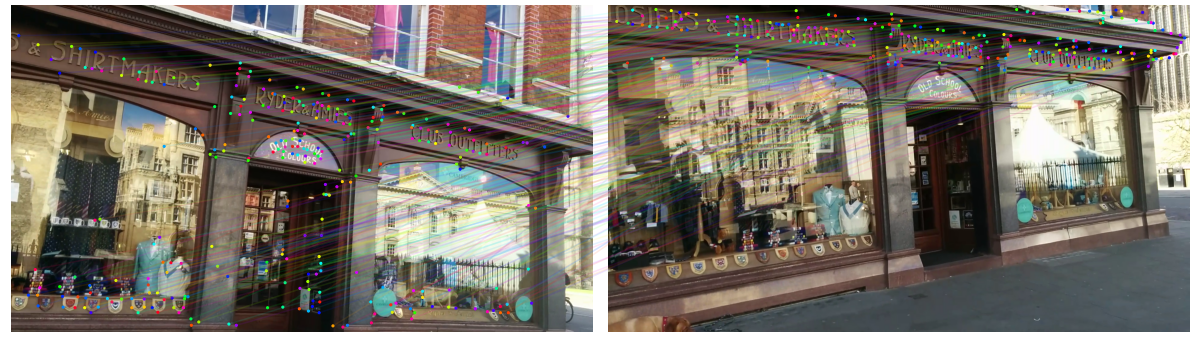}
  \includegraphics[width=\linewidth]{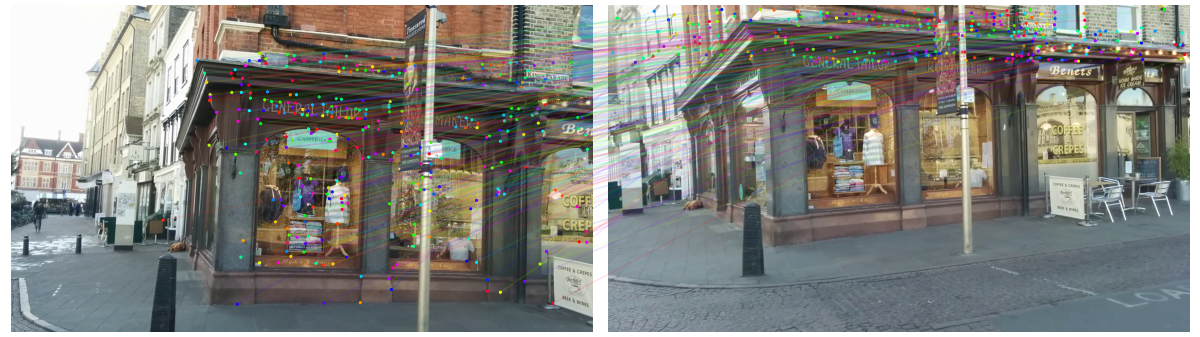}
  \caption{PQ4 + D-PQED}
\end{subfigure}%
\caption{\textbf{Qualitative results: PQ vs. D-PQED descriptors.} We evaluate both descriptors on the Aachen Day-Night and Cambridge Landmarks localization datasets~\cite{sattler2018benchmarking,Sattler2012BMVC,kendall2015posenet} and visualize local descriptors correspondences (in colors) produced by the SuperGlue matcher~\cite{sarlin2020superglue}. Both quantization techniques have a similar memory budget of 4MB. The proposed D-PQED layer provides more accurate correspondences leading to better localization performance.}\label{fig:qualitative_pq64_dpqed64_aachen_cambridge} \end{figure}

%% file: tbl/ablations.tex
\setlength{\tabcolsep}{10pt}
\begin{table}[t!]
\caption{\textbf{Ablation of Different Architecture Configuration.} We show localization performance on the Aachen Day-Night dataset for the baseline (PQ with the number of sub-vectors M=4) and different components of the proposed approach.} 
\centering
\scriptsize
\resizebox{\textwidth}{!}{
\begin{tabular}{lcccccc}  
\toprule
{} &  \multicolumn{6}{c}{Aachen Day-Night} \\
 & \multicolumn{3}{c}{Day-time queries} & \multicolumn{3}{c}{Night-time queries} \\
\midrule
Baseline: PQ (M=4) & 81.2 & 88.6 & 92.8 & 45.9 & 53.1 & 62.2 \\
+ Encoder ($\mathcal{E}$) & 84.6 & 93.0 & 96.8 &	68.4 & 80.6 & 89.8 \\
+ Decoder ($\mathcal{D}$) w/L2 loss & 86.0 & 93.3 & 97.2	 & 70.4 & 80.6 & 89.8 \\
+ Decoder ($\mathcal{D}$) w/Triplet loss & 86.8 & 93.1 & 97.8	 & 79.6 & 88.8 & 95.9 \\
+ D-PQED (Symmetric) & 86.4 & 93.6 & 97.7 &	74.5 & 86.7 & 95.9 \\
+ D-PQED & \textbf{88.1} & \textbf{95.4} & \textbf{98.4} &	\textbf{81.6} & \textbf{89.8} & \textbf{98.0} \\
\bottomrule
\end{tabular}
}
\label{subtab:hloc-vs-pq1}
\end{table}

%% file: sections/conclusion.tex
\section{Conclusion}\label{sec:conclusion}

In this work, we presented a differentiable product quantization method to reduce the memory necessary to store the visual descriptors while preserving their matching capability, hence enabling efficient localization.
The high accuracy matching of the memory-efficient descriptors is facilitated by the use of an unquantizer network that recovers the information of the original descriptors. 
The network is trained to not only recover the original descriptor signal but it is also encouraged to preserve the descriptor's metric space. Our results illustrate the efficiency of the proposed quantization and how it can seamlessly be integrated with further map sampling to comply with extremely low memory requirements with little or reasonable drop in localization performance.

%% file: supplementary/intro.tex
\title{Supplementary Material for \\ Differentiable Product Quantization for Memory Efficient Camera Relocalization}

\author{}
\institute{}
\authorrunning{Z. Laskar et al.}

\maketitle

In this \textbf{supplementary material}, we first discuss architectural and implementation details in~\secref{sec:implementation}. Next, we provide additional ablation studies of the proposed quantization approach for different local image descriptor matchers in~\secref{sec:ablation} and report additional quantitative and qualitative results in~\secref{sec:results}. Finally, we discuss potential negative impact of this work in~\secref{sec:impact}.

%% file: supplementary/implementation.tex
\section{Implementation Details}\label{sec:implementation}

For our single MLP decoder architecture, we use an 2-layer MLP with hidden dimension 256. The ReLU activation is used for the decoder. For experiments on the Aachen Day-Night dataset~\cite{sattler2018benchmarking,Sattler2012BMVC}, we extract local image descriptors from database images and create training and validation splits. We optimize our randomly initialized model for 30 epochs with batch size 1000 which takes about 7 hours on a single NVIDIA RTX2080Ti GPU. The model has 131072 parameters in total which is less than 0.5MB of memory. For the Cambridge Landmarks dataset~\cite{kendall2015posenet} and the 7 Scenes dataset~\cite{shotton2013scene}, we use our model pre-trained on Aachen. However, instead of fine-tuning the whole model on corresponding training and validation splits, we optimize only LoRA~\cite{hu2021lora} weights of 2048 parameters resulting in an overhead of 8KB memory.

%% file: supplementary/ablation.tex
\section{Ablation}\label{sec:ablation}
In this section, we first conduct several ablation studies to verify the effectiveness of the proposed approach, including different loss functions in~\secref{supp_ssec:ablation_loss_function} and different local image descriptor matchers in~\secref{supp_ssec:ablation_matchers}.

\subsection{Ablation of Different Loss Functions}\label{supp_ssec:ablation_loss_function}
To evaluate the effectiveness of our differential product quantization layer, we conduct ablation studies on the Aachen Day-Night dataset with three different loss functions. Specifically, we consider: a) \textbf{combined Triplet loss} (as discussed in the main part); b) vanilla \textbf{L2 loss}~\cite{Yang_2022_CVPR}; c) the \textbf{N-pair loss}~\cite{kihyuk_npair_loss}. Unlike traditional loss functions that typically compare a single pair of examples, \eg, contrastive loss or triplet loss, N-pair loss extends this idea to consider multiple negative examples simultaneously for each positive pair. This approach aims to improve the quality of the learned embeddings by enforcing a wider margin between the positive pairs and the multiple negative examples. Given a batch with pairs of examples, each pair consists of an anchor $x$ and a positive example $x^+$, and for each pair, there are $N-1$ negative examples $x_1^-, x_2^-, ... x_{N-1}^-$. The N-pair loss aims to make the anchor closer to its positive example than to any of the $N-1$ negatives and can be defined as follows:

\begin{equation}
L = \frac{1}{N} \sum_{i=1}^{N} \log \left[ 1 + \sum_{j=1}^{N-1} \exp \left( f(x_i) \cdot f(x^-_{ij}) - f(x_i) \cdot f(x^+_i) \right) \right]
\end{equation}

\noindent Here, $f(x_i)$ is the embedding function for the anchor, $f(x^+_i)$ is the embedding of the positive example associated with the anchor $x_i$, and $f(x^-_{ij})$ represents the embedding of the $j$th negative example for the anchor $x_i$. The outer sum iterates over each of the $N$ anchor-positive pairs in the batch, and the inner sum iterates over the $N-1$ negative examples associated with each anchor-positive pair.

We observe that using the proposed combined triplet loss function leads to the best results indicating that metric learning losses are beneficial compared to the reconstruction loss L2~\cite{Yang_2022_CVPR} for the camera relocalization task. The quantitative results are presented in~\tabref{supp_tab:ablation-losses}.

\setlength{\tabcolsep}{8pt}
\begin{table}[t!]
\caption{\textbf{Ablations of different loss functions}. Using the proposed triplet loss improves localization quality, with the percentages of accurate localized queries at three thresholds $(0.25/2), (0.5/5), (5,10) \text{m}/^o$.} 
\centering
\scriptsize
\resizebox{\textwidth}{!}{
\begin{tabular}{lcccccc}  
\toprule
{} &  \multicolumn{6}{c}{Aachen Day-Night} \\
 & \multicolumn{3}{c}{Day-time queries} & \multicolumn{3}{c}{Night-time queries} \\
\midrule
PQ4 & 81.2 & 88.6 & 92.8	& 45.9 & 53.1 & 62.2 \\
D-PQED + L2 & 86.0 & 93.3 & 97.2	 & 70.4 & 80.6 & 89.8 \\
D-PQED + N-pair (N=10) & 86.0 & 93.4 & 97.6	 & 76.5 & 86.7 & 95.4 \\
D-PQED + N-pair (N=20) & 84.7 & 92.2 & 96.6	 & 73.5 & 83.7 & 94.9 \\
D-PQED + Triplet loss (m=0.5) & 86.3 & 93.3 & 97.9 & 76.5 & 90.8 & 95.9 \\
D-PQED + Triplet loss (m=0.3) & 86.8 & 93.8 & 97.8	& 73.5 & 86.7 & 96.9 \\
D-PQED + Triplet loss (m=0.9) & 86.8 & 93.1 & 97.8 & 79.6 & 88.8 & 95.9 \\
\bottomrule
\end{tabular}%
}
\label{supp_tab:ablation-losses}
\end{table}%

\begin{table}[t!]
\caption{\textbf{Ablation of different matchers.} We observe that using the nearest neighbor (NN) mutual matcher leads to a significant drop for both hard (PQ) and differential D-PQED quantization techniques. }
\centering
\scriptsize
\resizebox{\textwidth}{!}{
\begin{tabular}{clccccccc}  
\toprule
{} & {} & {} & \multicolumn{6}{c}{Aachen Day-Night} \\
{} &  & MB & \multicolumn{3}{c}{Day-time queries} & \multicolumn{3}{c}{Night-time queries} \\
\midrule
\multirow{3}{*}{\rotatebox[origin=c]{90}{NN}} & HLoc-avg. & 618 & 85.3 & 91.3 & 95.6 & 74.5 & 84.7 & 92.9 \\
 & PQ4 & 4 & 72.3 & 80.0 & 86.4 & 38.8 & 40.8 & 52.0 \\
 & D-PQED & 4 & 70.0 & 77.1 & 82.0 &	38.8 & 40.8 & 44.9 \\
 \midrule
\multirow{3}{*}{\rotatebox[origin=c]{90}{\parbox{1cm}{\centering SGlue\\~\cite{sarlin2020superglue}}}} & HLoc-avg. & 618 & 88.5 & 95.8 & 98.8 & 84.7 & 93.9 & 100.0 \\
 & PQ4 & 4 & 81.2 & 88.6 & 92.8	& 45.9 & 53.1 & 62.2 \\
 & D-PQED & 4 & 88.1 & 95.4 & 98.4 &	81.6 & 89.8 & 98.0 \\
 \addlinespace[0.12cm]
\bottomrule
\end{tabular}%
}
\label{supp_tab:ablation-matchers}
\end{table}

\subsection{Ablation of Different Matchers }\label{supp_ssec:ablation_matchers}
To further analyze the localization performance of the proposed differentiable quantization layer, we further tested our model with different local descriptor matchers. Specifically, we consider the mutual Nearest Neighbor (NN) matcher and SuperGlue~\cite{sarlin2020superglue}. The results presented in~\tabref{supp_tab:ablation-matchers} demonstrate that neither PQ nor its differentiable counterpart (D-PQED) is compatible with the NN matcher showing a significant drop in localization accuracy. This is quite an interesting observation. We hypothesize that this behavior could be explained by SuperGlue's ability to leverage the global context. Given its learning-based nature, SuperGlue is trained to handle real-world imperfections in descriptors and can still perform well even when descriptors are not perfectly discriminative on their own. In contrast, the performance of the NN matcher heavily depends on the discriminative power of individual descriptors and it does not consider the overall geometry or the relationship between keypoints making it prone to mismatches in challenging scenarios, \ie localizing night-time queries (\cf the right-most part of~\tabref{supp_tab:ablation-matchers}).

%% file: supplementary/results.tex
\input{tbl/7s_detail}

\input{tbl/cam_details}

\begin{figure}[h!]
\centering
\begin{subfigure}{.49\textwidth}
  \centering
  \includegraphics[width=\linewidth]{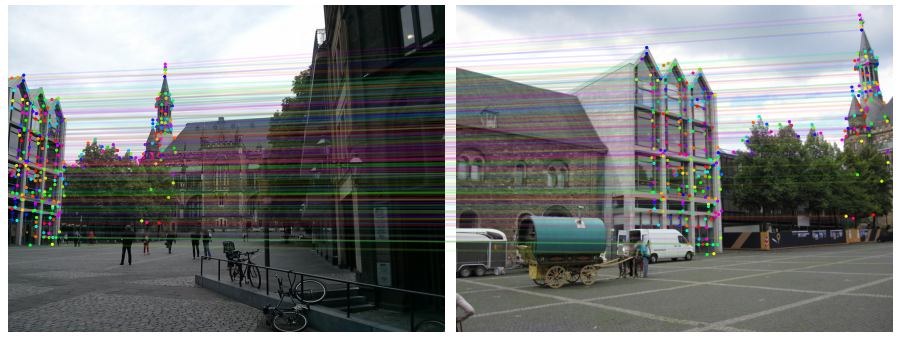}
  \includegraphics[width=\linewidth]{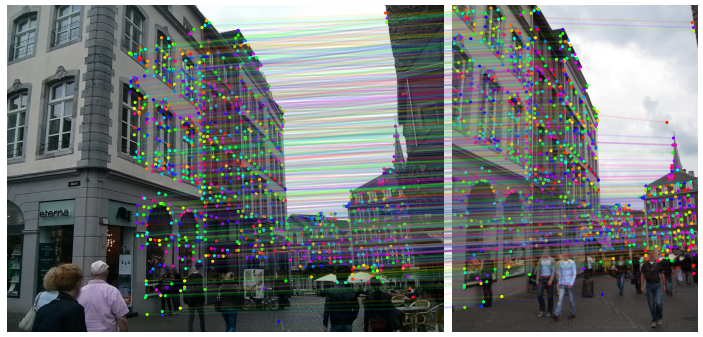}
  \includegraphics[width=\linewidth]{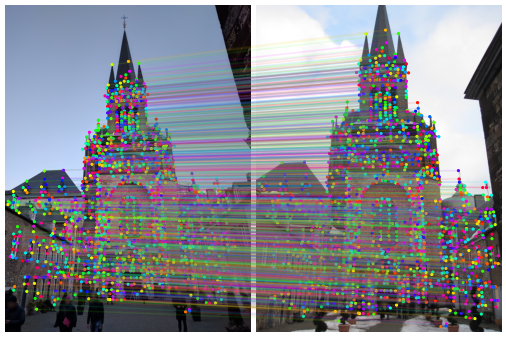}
  \includegraphics[width=\linewidth]{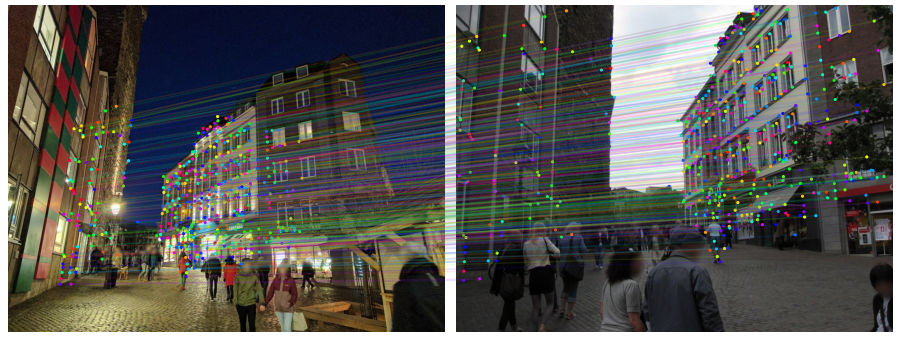}
  \includegraphics[width=\linewidth]{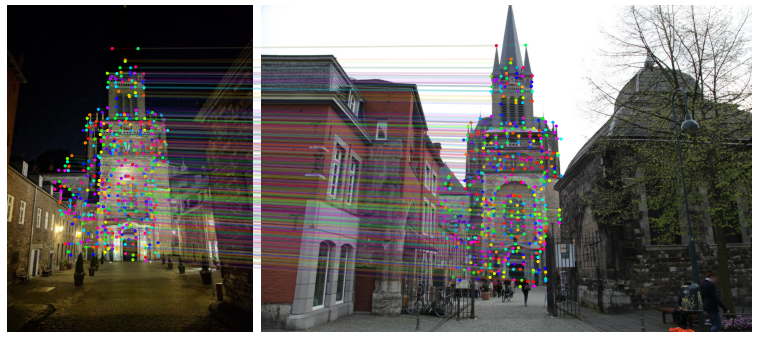}
  \caption{SuperPoint+SuperGlue (618MB)}
\end{subfigure}%
\hfill
\begin{subfigure}{.49\textwidth}
  \centering
  \includegraphics[width=\linewidth]{supplementary/fig/matches/matches_pq64_day_1.png}
  \includegraphics[width=\linewidth]{supplementary/fig/matches/matches_pq64_day_2.png}
  \includegraphics[width=\linewidth]{supplementary/fig/matches/matches_pq64_day_5.png}
  \includegraphics[width=\linewidth]{supplementary/fig/matches/matches_pq64_night_1.png}
  \includegraphics[width=\linewidth]{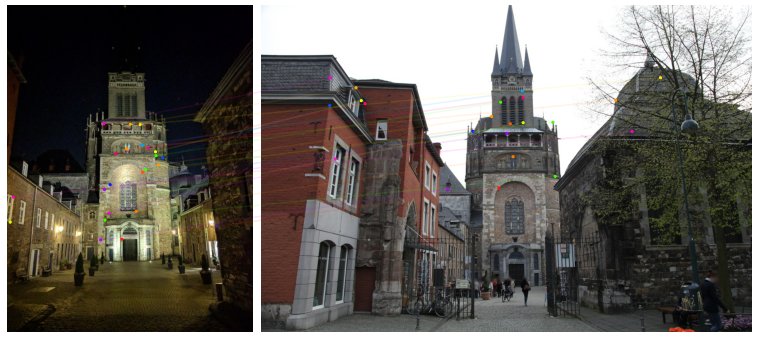}
  \caption{PQ4 (4MB)}
\end{subfigure}%
\caption{\textbf{Qualitative results on Aachen: SuperPoint vs. PQ4 descriptors.} We evaluate both descriptors on the Aachen Day-Night localization dataset~\cite{sattler2018benchmarking,Sattler2012BMVC} and visualize local descriptors correspondences produced by the SuperGlue matcher~\cite{sarlin2020superglue}. Although having significantly low memory consumption, the hard PQ descriptors struggle to provide accurate correspondences leading to weak localization performance. Please zoom in to see the details.}\label{sup_fig:superpoint_pq64_aachen}
\end{figure}

\begin{figure}[h!]
\centering
\begin{subfigure}{.49\textwidth}
  \centering
  \includegraphics[width=\linewidth]{supplementary/fig/matches/matches_day_1.png}
  \includegraphics[width=\linewidth]{supplementary/fig/matches/matches_day_2.png}
  \includegraphics[width=\linewidth]{supplementary/fig/matches/matches_day_5.png}
  \includegraphics[width=\linewidth]{supplementary/fig/matches/matches_night_1.png}
  \includegraphics[width=\linewidth]{supplementary/fig/matches/matches_night_2.png}
  \caption{SuperPoint+SuperGlue (618MB)}
\end{subfigure}%
\hfill
\begin{subfigure}{.49\textwidth}
  \centering
  \includegraphics[width=\linewidth]{supplementary/fig/matches/matches_pq64_e2e_day_1.png}
  \includegraphics[width=\linewidth]{supplementary/fig/matches/matches_pq64_e2e_day_2.png}
  \includegraphics[width=\linewidth]{supplementary/fig/matches/matches_pq64_e2e_day_5.png}
  \includegraphics[width=\linewidth]{supplementary/fig/matches/matches_pq64_e2e_night_1.png}
  \includegraphics[width=\linewidth]{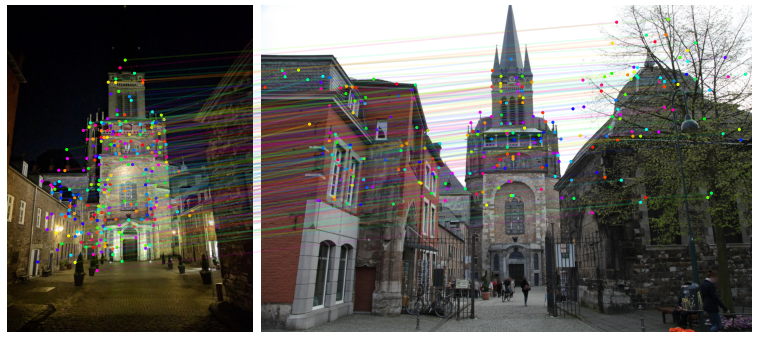}
  \caption{PQ4+D-PQED (4MB)}
\end{subfigure}%
\caption{\textbf{Qualitative results on Aachen: SuperPoint vs. proposed D-PQED descriptors.} Similar to~\figref{sup_fig:superpoint_pq64_aachen}, both descriptors are evaluated on the Aachen Day-Night localization dataset and the SuperGlue matcher~\cite{sarlin2020superglue} is used to establish 2D-2D inliers. In contrast to its non-differentiable counterpart (\cf~\figref{sup_fig:superpoint_pq64_aachen}), the proposed D-PQED layer produces better correspondences. }\label{sup_fig:superpoint_nipq64_aachen}
\end{figure}

\section{Additional Results} \label{sec:results}
\begin{figure}[h!]
\centering

\begin{subfigure}{.49\textwidth}
  \centering
  \includegraphics[width=\linewidth]{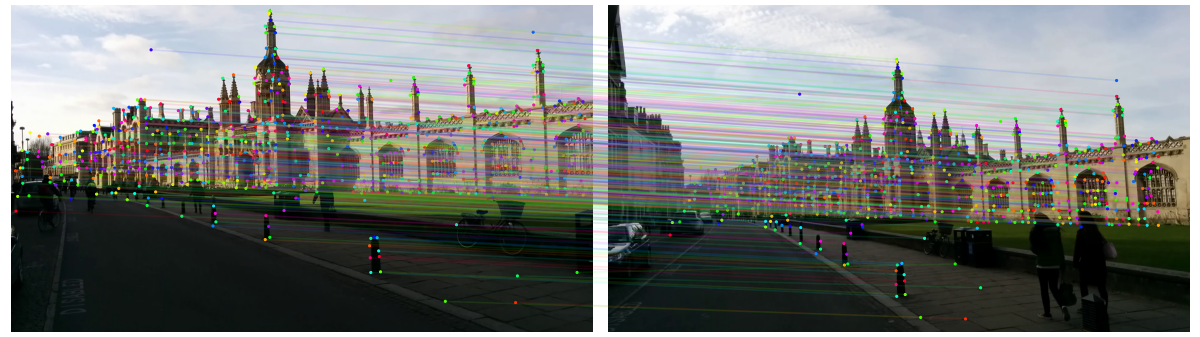}
  \includegraphics[width=\linewidth]{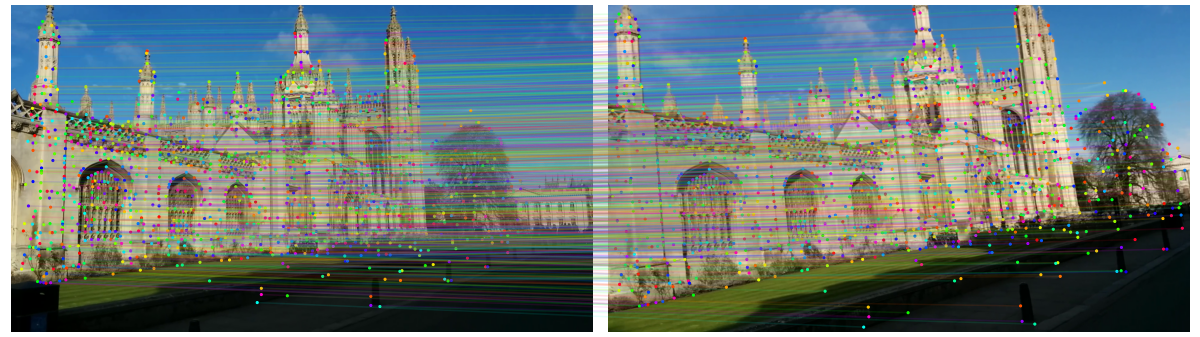}
  \includegraphics[width=\linewidth]{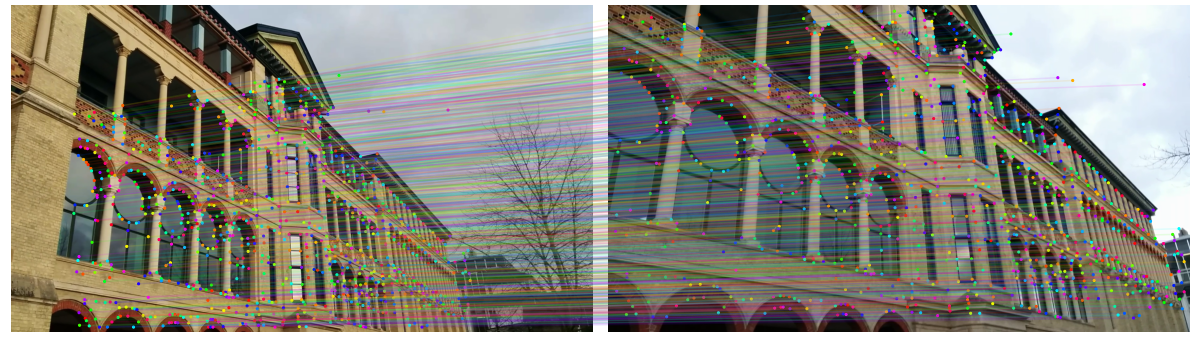}
  \includegraphics[width=\linewidth]{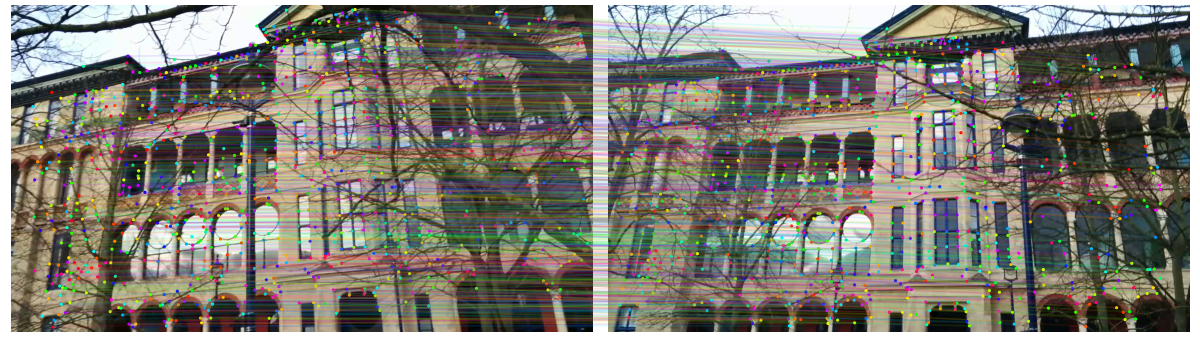}
  \includegraphics[width=\linewidth]{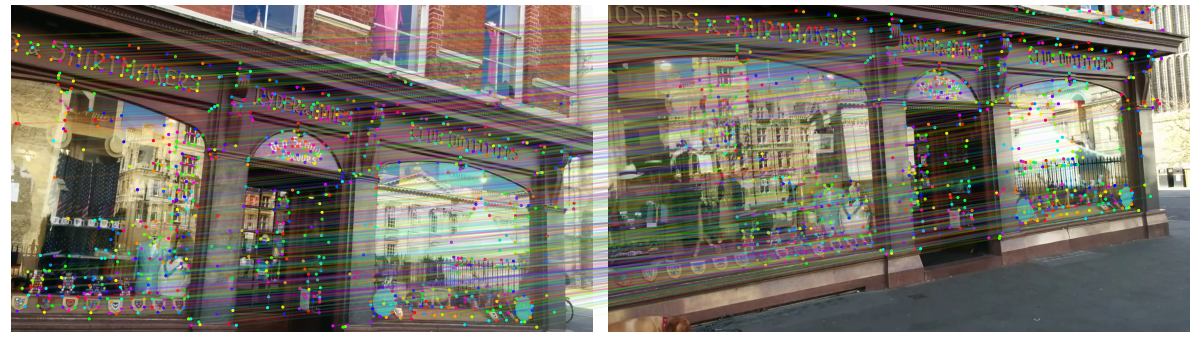}
  \includegraphics[width=\linewidth]{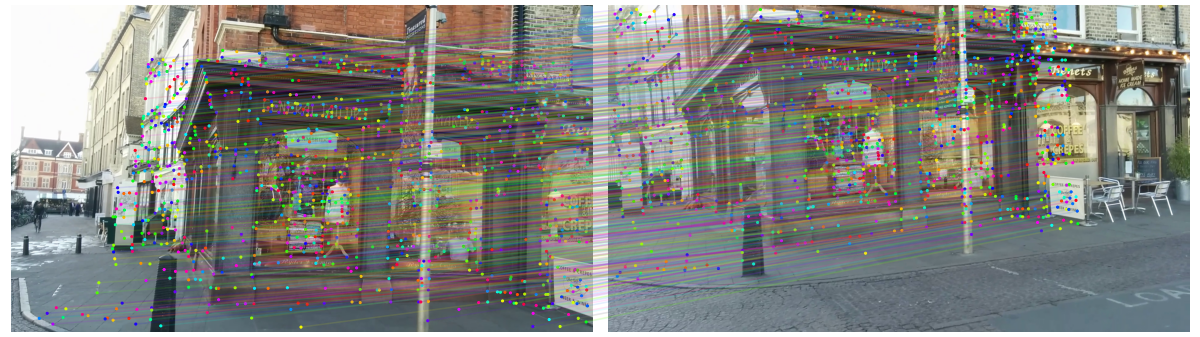}
  \caption{SuperPoint+SuperGlue (618MB)}
\end{subfigure}%
\hfill
\begin{subfigure}{.49\textwidth}
  \centering
  \includegraphics[width=\linewidth]{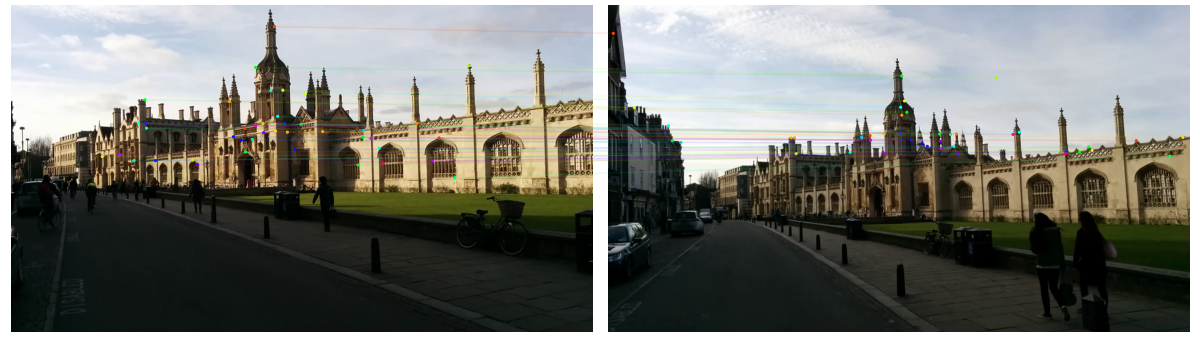}
  \includegraphics[width=\linewidth]{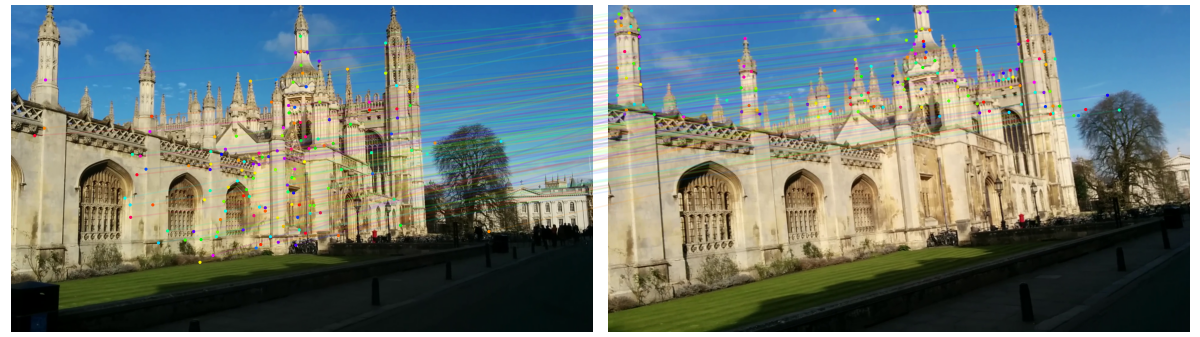}
  \includegraphics[width=\linewidth]{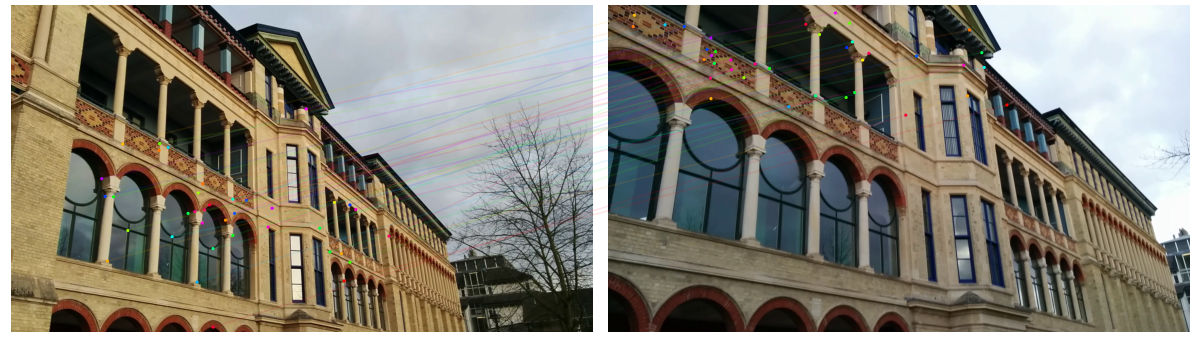}
  \includegraphics[width=\linewidth]{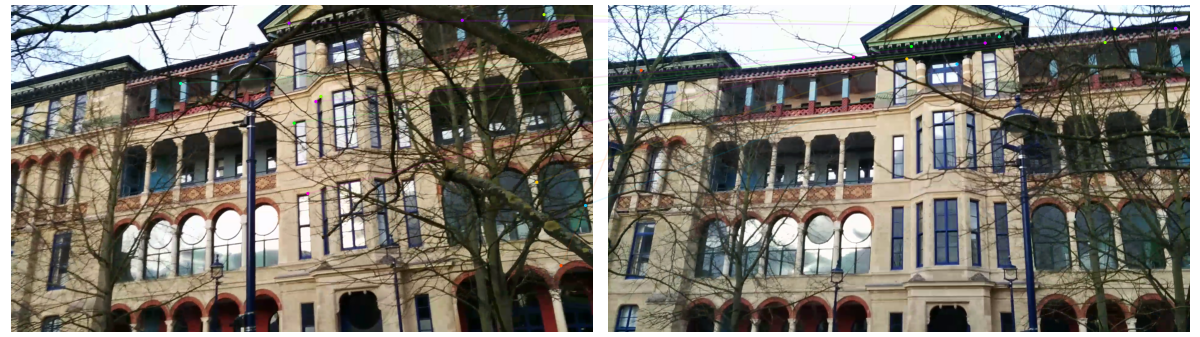}
  \includegraphics[width=\linewidth]{supplementary/fig/matches/cambridge/matches_pq64_5.png}
  \includegraphics[width=\linewidth]{supplementary/fig/matches/cambridge/matches_pq64_6.png}
  \caption{PQ4 (4MB)}
\end{subfigure}%
\caption{\textbf{Qualitative results on Cambridge: SuperPoint vs. PQ4 descriptors.} Similar to~\figref{sup_fig:superpoint_pq64_aachen} and~\figref{sup_fig:superpoint_nipq64_aachen}, we evaluate both descriptors on the Cambridge Landmarks localization dataset and use the SuperGlue matcher~\cite{sarlin2020superglue} to establish a 2D-2D set of inliers. The hard PQ descriptors fail to short to produce reliable correspondences. Please zoom in to see the details.}\label{sup_fig:superpoint_pq64_cambridge}
\end{figure}

\begin{figure}[h!]
\centering

\begin{subfigure}{.49\textwidth}
  \centering
  \includegraphics[width=\linewidth]{supplementary/fig/matches/cambridge/matches_1.png}
  \includegraphics[width=\linewidth]{supplementary/fig/matches/cambridge/matches_2.png}
  \includegraphics[width=\linewidth]{supplementary/fig/matches/cambridge/matches_3.png}
  \includegraphics[width=\linewidth]{supplementary/fig/matches/cambridge/matches_4.png}
  \includegraphics[width=\linewidth]{supplementary/fig/matches/cambridge/matches_5.png}
  \includegraphics[width=\linewidth]{supplementary/fig/matches/cambridge/matches_6.png}
  \caption{SuperPoint+SuperGlue (618MB)}
\end{subfigure}%
\hfill
\begin{subfigure}{.49\textwidth}
  \centering
  \includegraphics[width=\linewidth]{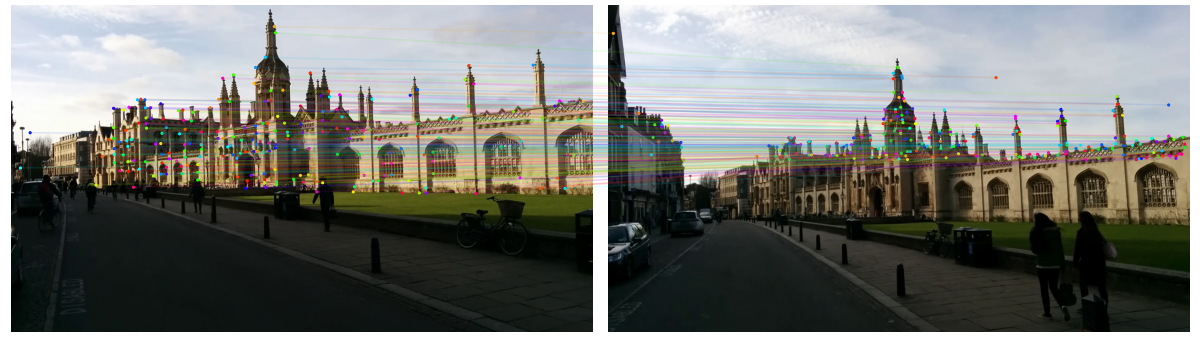}
  \includegraphics[width=\linewidth]{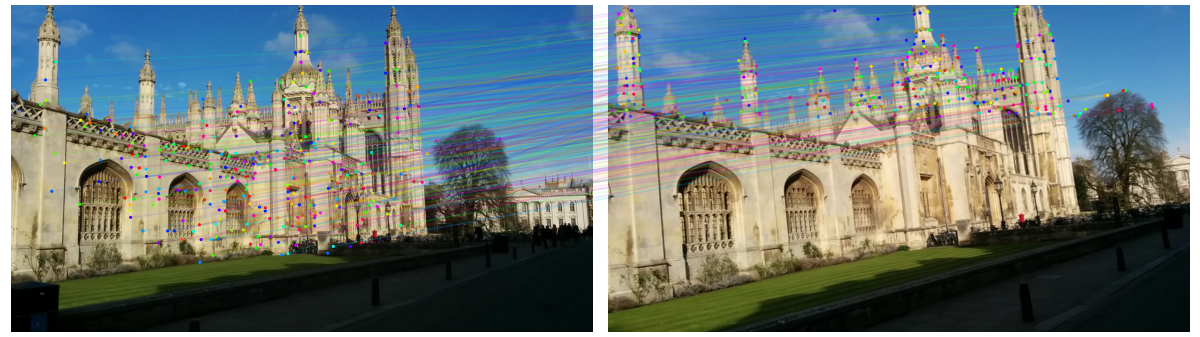}
  \includegraphics[width=\linewidth]{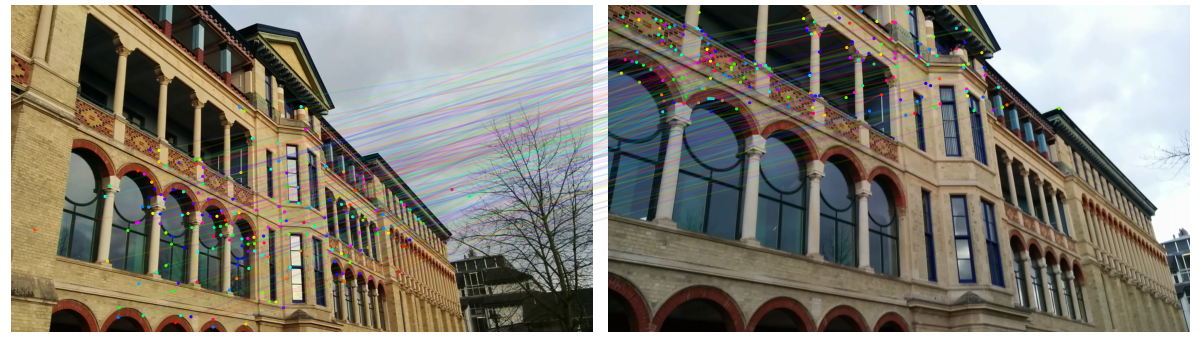}
  \includegraphics[width=\linewidth]{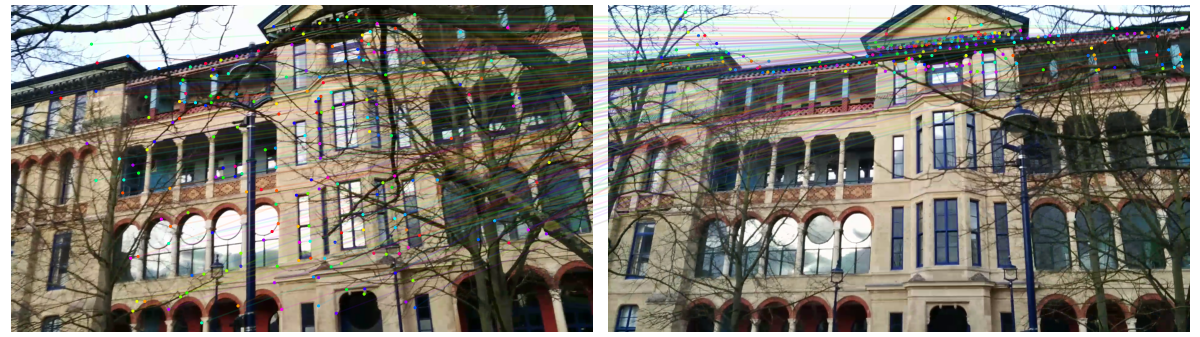}
  \includegraphics[width=\linewidth]{supplementary/fig/matches/cambridge/matches_pq64_nipq_5.png}
  \includegraphics[width=\linewidth]{supplementary/fig/matches/cambridge/matches_pq64_nipq_6.png}
  \caption{PQ4+D-PQED (4MB)}
\end{subfigure}%
\caption{\textbf{Qualitative results on Cambridge: SuperPoint vs. proposed D-PQED descriptors.} As for Aachen (\cf~\figref{sup_fig:superpoint_pq64_aachen} and~\figref{sup_fig:superpoint_nipq64_aachen}), the proposed D-PQED layer produces better correspondences compared to PQ4.}\label{sup_fig:superpoint_nipq64_cambridge}
\end{figure}

In this section, we provide more qualitative and quantitative results for three datasets: Aachen Day-Night~\cite{sattler2018benchmarking,Sattler2012BMVC}, 7Scenes, and Cambridge Landmarks~\cite{kendall2015posenet}. We show more visualizations in~\figref{sup_fig:superpoint_pq64_aachen},~\figref{sup_fig:superpoint_nipq64_aachen},~\figref{sup_fig:superpoint_pq64_cambridge},~\figref{sup_fig:superpoint_nipq64_cambridge}. Specifically, we evaluate SuperPoint~\cite{detone2018superpoint}, PQ64 and the proposed differentiable D-PQED descriptors on Aachen and Cambridge. The SuperGlue~\cite{sarlin2020superglue} matcher is used to establish correspondences between keypoints extracted from a query and its corresponding database image. Compared to the hard PQ descriptors, the proposed approach provides better correspondences leading to more accurate localization performance with a smaller memory budget compared to SuperPoint. The localization results on the 7Scenes dataset and Cambridge Landmark dataset are provided in~\tabref{supp_tab:7s_detailed} and~\tabref{supp_tab:cambridge_detailed}, respectively.

%% file: tbl/7s_detail.tex
\setlength{\tabcolsep}{4pt}
\begin{table}[t!]
\caption{ \textbf{Results on 7Scenes.} We report storage requirements (MB), median translation (m), and rotation ($^\circ$) errors and the average accuracy (\%) on 7Scenes.}
\small
\centering 
\resizebox{\textwidth}{!}{
\begin{tabular}{lrcccccccc} %
\toprule
		&  & \multicolumn{8}{c}{7Scenes}  \\

         & MB & chess & fire & heads & office & pumpkin & redkitchen & stairs & avg \\
\midrule
HLoc~\cite{sarlin2019coarse, sarlin2020superglue} & 9688 & 0.02/0.80/0.94 & 0.02/0.77/0.97 & 0.01/0.74/0.99 & 0.03/0.83/0.84 & 0.04/1.05/0.62 & 0.03/1.12/0.72 & 0.05/1.26/0.54  & 0.03/0.93/0.81
 \\
HLoc-avg & 616 & 0.02/0.79/0.94 & 0.02/0.77/0.96 & 0.01/0.73/0.99 & 0.03/0.84/0.83 & 0.04/1.05/0.63 & 0.03/1.12/0.72 & 0.04/1.19/0.57  & 0.03/0.94/0.80
 \\
\midrule
PQ32 & 37 & 0.02/0.80/0.94 & 0.02/0.77/0.97 & 0.01/0.73/0.99 & 0.03/0.84/0.83 & 0.04/1.07/0.60 & 0.04/1.07/0.60 & 0.05/1.19/0.51  & 0.03/0.93/0.78
 \\
PQ16 & 19 & 0.02/0.81/0.94 & 0.02/0.76/0.97 & 0.01/0.73/0.99 & 0.01/0.73/0.99 & 0.04/1.09/0.60 & 0.03/1.13/0.72 & 0.05/1.33/0.46  & 0.03/0.94/0.81
 \\
PQ8 & 9 & 0.02/0.80/0.94 & 0.02/0.76/0.97 & 0.01/0.75/0.99 & 0.01/0.75/0.99 & 0.04/1.11/0.60 & 0.03/1.14/0.71 & 0.06/1.45/0.44  & 0.03/0.97/0.80
 \\
PQ4 & 5 & 0.02/0.82/0.93 & 0.02/0.79/0.96 & 0.01/0.76/0.99 & 0.03/0.84/0.83 & 0.04/1.12/0.60 & 0.03/1.16/0.71 & 0.06/1.49/0.41  & 0.03/1.00/0.78
 \\
PQ2 & 2 & 0.02/0.81/0.93 & 0.02/0.81/0.93 & 0.01/0.77/0.98 & 0.03/0.86/0.82 & 0.04/1.16/0.58 & 0.04/1.16/0.58 & 0.06/1.62/0.40  & 0.03/1.03/0.75
 \\
\midrule
PQ32 + $\Gamma_{.008}$ & 0.3 & 0.05/1.60/0.53 & 0.04/1.39/0.65 & 0.02/1.17/0.7 & 0.07/1.89/0.36 & 0.09/2.29/0.27 & 0.07/2.01/0.33 & 0.18/5.76/0.23 & 0.07/2.30/0.44
 \\
+ D-PQED & 0.3 & 0.04/1.50/0.56 & 0.04/1.34/0.69 & 0.02/1.13/0.71 & 0.07/1.80/0.37 & 0.08/2.07/0.32 & 0.07/1.95/0.34 & 0.15/4.48/0.26 & 0.07/2.04/0.46
 \\
PQ4 + $\Gamma_{.064}$ & 0.3 & 0.03/0.99/0.87 & 0.02/0.90/0.88 & 0.01/0.79/0.94 & 0.03/0.99/0.73 & 0.05/1.39/0.47 & 0.04/1.33/0.61 & 0.06/1.65/0.40 & 0.04/1.15/0.70
 \\
+ D-PQED & 0.3 & 0.03/0.97/0.89 & 0.02/0.92/0.87 & 0.01/0.80/0.94 & 0.03/0.97/0.74 & 0.05/1.25/0.49 & 0.04/1.28/0.63 & 0.05/1.36/0.52 & 0.03/1.08/0.73
 \\
PQ2 + $\Gamma_{.128}$ & 0.3 & 0.03/0.92/0.91 & 0.02/0.89/0.90 & 0.01/0.81/0.95 & 0.03/0.93/0.77 & 0.05/1.27/0.52 & 0.04/1.36/0.63 & 0.06/1.51/0.42 & 0.03/1.09/0.73
 \\
+ D-PQED & 0.3 & 0.03/0.87/0.92 & 0.02/0.99/0.87 & 0.01/0.82/0.94 & 0.03/0.91/0.78 & 0.05/1.18/0.54 & 0.04/1.30/0.65 & 0.06/1.48/0.44 & 0.03/1.08/0.73 \\

\bottomrule

\end{tabular}
}
\label{supp_tab:7s_detailed}
\end{table}

%% file: tbl/cam_details.tex
\setlength{\tabcolsep}{4pt}
\begin{table}[t!]
\caption{\textbf{Results on Cambridge Dataset.} We report storage requirements (MB), median translation (m), and rotation ($^\circ$) errors. } \label{supp_tab:cambridge_detailed}
\small
\centering 
\resizebox{\textwidth}{!}{
\begin{tabular}{lrcccccc} %
\toprule
		& & \multicolumn{6}{c}{Cambridge}  \\

         & MB & GreatCourt & KingsCollege & OldHospital & ShopFacade & StMarysChurch & Avg \\
\midrule

HLoc~\cite{sarlin2019coarse, sarlin2020superglue} & 5440 & 0.18/0.11 & 0.11/0.21 & 0.15/0.31 & 0.04/0.20 & 0.07/0.22 & 0.11/0.21
 \\
HLoc-avg & 518 & 0.18/0.11 & 0.11/0.21 & 0.15/0.32 & 0.04/0.19 & 0.07/0.22 & 0.11/0.21
 \\



\midrule
 
PQ32 & 32 & 0.18/0.11 & 0.12/0.21 & 0.14/0.29 & 0.04/0.20 & 0.07/0.22 & 0.11/0.21
 \\
PQ16 & 16 & 0.19/0.11 & 0.12/0.21 & 0.15/0.29 & 0.05/0.23 & 0.07/0.23 & 0.12/0.21
 \\
PQ8 & 8 & 0.20/0.11 & 0.12/0.23 & 0.16/0.31 & 0.04/0.24 & 0.07/0.23 & 0.12/0.22
 \\
PQ4 & 4 & 0.22/0.12 & 0.13/0.22 & 0.17/0.30 & 0.05/0.24 & 0.08/0.24 & 0.13/0.23
 \\
PQ2 & 2 & 0.24/0.13 & 0.12/0.22 & 0.16/0.32 & 0.05/0.25 & 0.08/0.25 & 0.13/0.23
 \\

\midrule
 
PQ32 + $\Gamma_{.008}$ & 0.3 & 0.63/0.25 & 0.25/0.38 & 0.37/0.57 & 0.08/0.36 & 0.15/0.47 & 0.29/0.41
 \\
+ D-PQED & 0.3 & 0.46/0.19 & 0.25/0.37 & 0.28/0.42 & 0.06/0.34 & 0.13/0.42 & 0.24/0.35
 \\
PQ4 + $\Gamma_{.064}$ & 0.3 & 0.44/0.22 & 0.18/0.31 & 0.25/0.42 & 0.06/0.31 & 0.11/0.35 & 0.21/0.32
 \\
+ D-PQED & 0.3 & 0.25/0.13 & 0.16/0.28 & 0.16/0.29 & 0.06/0.28 & 0.09/0.27 & 0.14/0.25
 \\
PQ2 + $\Gamma_{.0128}$ & 0.3 & 0.41/0.21 & 0.16/0.28 & 0.20/0.31 & 0.06/0.30 & 0.10/0.32 & 0.18/0.28
 \\
+ D-PQED & 0.3 & 0.22/0.12 & 0.15/0.25 & 0.15/0.29 & 0.05/0.26 & 0.08/0.26 & 0.13/0.24 \\

\bottomrule

\end{tabular}
}

\end{table}

%% file: supplementary/impact.tex
\section{Societal Impact}\label{sec:impact}
Our method can efficiently relocalize the camera with minimal memory requirement, which can be applied in augmented reality, robotics, and autonomous driving. However, it's crucial to also acknowledge potential negative impacts. Specifically, the widespread adoption of advanced localization technologies could lead to heightened surveillance and privacy concerns. Additionally, the reliance on such technologies in critical systems like autonomous driving raises safety and reliability issues; any failure or manipulation of the localization process could have dire consequences. Ethical considerations around the deployment of these technologies must be carefully weighed against their benefits, ensuring that safeguards are in place to prevent misuse and that efforts are made to minimize inequalities in access and impact.